\documentclass[10pt,twocolumn,letterpaper]{article}
\usepackage[pagenumbers]{cvpr}
\usepackage{graphicx} \graphicspath{{figures/}}
\usepackage{amsmath,amssymb,mathabx,mathtools,amsthm,nicefrac}
\usepackage{algorithmic}
\usepackage[linesnumbered,ruled,vlined]{algorithm2e}
\usepackage{acronym}
\definecolor{cvprblue}{rgb}{0.21,0.49,0.74}\usepackage[pagebackref,breaklinks,colorlinks,allcolors=cvprblue]{hyperref}
\usepackage{balance}
\usepackage{xspace}
\usepackage{setspace}
\usepackage[skip=3pt,font=small]{subcaption}
\usepackage[skip=3pt,font=small]{caption}
\usepackage[dvipsnames,svgnames,x11names,table]{xcolor}
\usepackage[capitalize]{cleveref}
\usepackage{booktabs,tabularx,colortbl,multirow,array,makecell,tabularray}
\usepackage{enumitem}
\usepackage[misc]{ifsym}
\usepackage{siunitx}

\makeatletter
\DeclareRobustCommand\onedot{\futurelet\@let@token\@onedot}
\def\@onedot{\ifx\@let@token.\else.\null\fi\xspace}
\def\eg{\emph{e.g}\onedot} 

\def\ie{\emph{i.e}\onedot}

\def\vs{\emph{vs}\onedot}

\makeatother

\makeatletter
\renewcommand{\paragraph}{%
  \@startsection{paragraph}{4}%
  {\z@}{0ex \@plus 0ex \@minus 0ex}{-1em}%
  {\hskip\parindent\normalfont\normalsize\bfseries}%
}
\makeatother

\acrodef{akr}[AKR]{Augmented Kinematic Representation}
\acrodef{momao}[MoMAO]{Mobile Manipulation of Articulated Objects}
\acrodef{slam}[SLAM]{simultaneous localization and mapping}
\acrodef{tamp}[TAMP]{task and motion planning}
\acrodef{urdf}[URDF]{Unified Robot Description Format}
\acrodef{pt}[\emph{pt}]{parse tree}
\acrodef{iou}[IoU]{intersection over union}
\acrodef{map}[mAP]{mean average precision}
\acrodef{dof}[DoF]{Degree of Freedom}
\acrodef{ik}[IK]{Inverse Kinematics}
\acrodef{fk}[FK]{Forward Kinematics}
\acrodef{rl}[RL]{Reinforcement Learning}
\acrodef{il}[IL]{Imitation Learning}
\acrodef{vla}[VLA]{Vision-Language-Action}
\acrodef{gpu}[GPU]{Graphics Processing Unit}
\acrodef{esdf}[ESDF]{Euclidean Signed-distance Field}
\acrodef{sota}[SOTA]{State-of-the-Art}
\acrodef{tcp}[TCP]{Tool Center Point}
\acrodef{rgbd}[RGB-D]{Red, Green, Blue - Depth}
\acrodef{ap}[AP]{Affinity Propagation}
\newcommand{\dataset}{\texttt{AutoMoMa}\xspace}

\newcommand{\Call}[2]{\texttt{#1}(\textrm{#2})}

\def\paperID{*****}
\def\confName{CVPR}
\def\confYear{2026}

\title{Scalable Trajectory Generation for Whole-Body Mobile Manipulation\vspace{-18pt}}

\author{
    Yida Niu \textsuperscript{1,3,4,5}\footnotemark[1] \quad
    Xinhai Chang \textsuperscript{1,3,4,5,6}\footnotemark[1] \quad
    Xin Liu \textsuperscript{4}\footnotemark[1] \quad
    Ziyuan Jiao \textsuperscript{2,4}$^{\,\textrm{\Letter}}$ \quad
    Yixin Zhu \textsuperscript{3,4,5,7}$^{\,\textrm{\Letter}}$
    \vspace{1pt}\\
    \small \textsuperscript{1} Institute for AI, Peking University\quad
    \small \textsuperscript{2} Institute of Unmanned System, Beihang University\vspace{-1pt}\\
    \small \textsuperscript{3} School of Psychological and Cognitive Sciences, Peking University\quad
    \small \textsuperscript{4} State Key Laboratory of General Artificial Intelligence\vspace{-1pt}\\
    \small \textsuperscript{5} Beijing Key Laboratory of Behavior and Mental Health, Peking University\quad
    \small \textsuperscript{6} Yuanpei College, Peking University\vspace{-1pt}\\
    \small \textsuperscript{7} Embodied Intelligence Lab, PKU-Wuhan Institute for Artificial Intelligence\vspace{-1pt}\\
    \small \footnotemark[1]\;\;Equal contribution \quad 
    \small $\textrm{\Letter}$\,\,\texttt{yixin.zhu@pku.edu.cn,\,zyjiao@buaa.edu.cn} \quad
    \small \href{https://automoma.pages.dev/}{https://automoma.pages.dev/}
    \vspace{-18pt}%
}%

\begin{document}

\twocolumn[{%
    \renewcommand\twocolumn[1][]{#1}%
    \maketitle
    \begin{center}
        \centering
        \captionsetup{type=figure}
        \includegraphics[width=\linewidth]{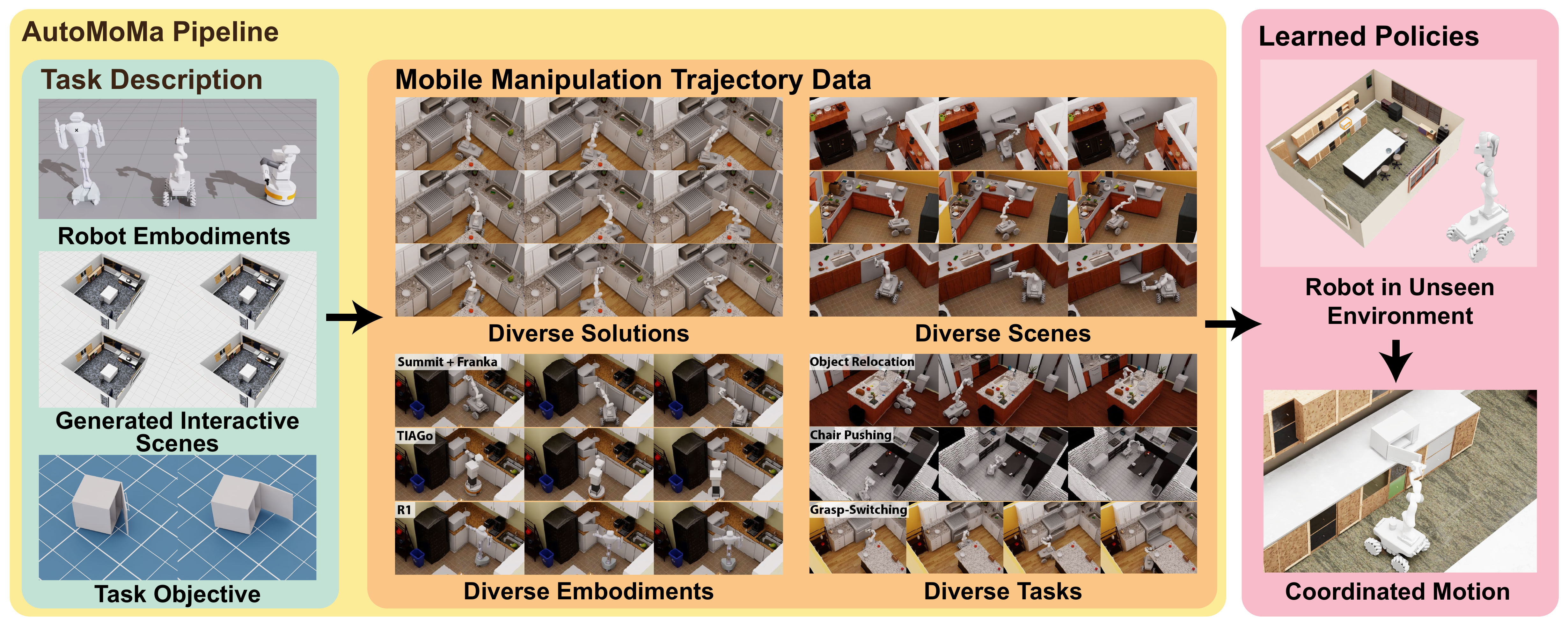} 
        \captionof{figure}{\textbf{Overview of the \dataset framework.} Coordinated mobile manipulation demands large-scale, physically valid trajectory data---a bottleneck that existing teleoperation and planning methods cannot overcome at scale. \dataset addresses this by unifying \ac{akr} modeling, which consolidates base, arm, and object kinematics into a single chain, with \acs{gpu}-accelerated trajectory optimization. Given diverse robot embodiments, interactive scenes, and task objectives as inputs (left), \dataset efficiently synthesizes over 500k trajectories exhibiting broad diversity across solutions, scenes, embodiments, and complex tasks such as grasp switching (center). This high-quality data enables the training of robust \ac{il} policies that generalize coordinated whole-body motion to unseen environments (right).}
        \label{fig:overview}
    \end{center}%
}]
\begin{abstract}
Robots deployed in unstructured environments must coordinate whole-body motion---simultaneously moving a mobile base and arm---to interact with the physical world.
This coupled mobility and dexterity yields a state space that grows combinatorially with scene and object diversity, demanding datasets far larger than those sufficient for fixed-base manipulation. Yet existing acquisition methods, including teleoperation~\cite{fu2024mobile} and planning~\cite{sleiman2023versatile,jiao2025integration}, are either labor-intensive or computationally prohibitive at scale.
The core bottleneck is the lack of a scalable pipeline for generating large-scale, physically valid, coordinated trajectory data across diverse embodiments and environments.
Here we introduce \dataset, a \acs{gpu}-accelerated framework that unifies \ac{akr} modeling, which consolidates base, arm, and object kinematics into a single chain, with parallelized trajectory optimization. \dataset achieves 5,000 episodes per \acs{gpu}-hour (over 80$\times$ faster than CPU-based baselines~\cite{zhang2025m}), producing a dataset of over 500k physically valid trajectories spanning 330 scenes, diverse articulated objects, and multiple robot embodiments.
Prior datasets were forced to compromise on scale, diversity, or kinematic fidelity~\cite{fu2024mobile,zhang2025m}; \dataset addresses all three simultaneously. Training downstream \ac{il} policies further reveals that even a single articulated-object task requires tens of thousands of demonstrations for \ac{sota} methods to reach $\approx$80\% success, confirming that data scarcity---not algorithmic limitations---has been the binding constraint.
\dataset thus bridges high-performance planning and reliable \acs{il}-based control, providing the infrastructure previously missing for coordinated mobile manipulation research.
By making large-scale, kinematically valid training data practical, \dataset showcases generalizable whole-body robot policies capable of operating in the diverse, unstructured settings of the real world.
\end{abstract}

\section{Introduction}

Whole-body mobile manipulation, which requires coordinated control of the mobile base and the arm, is fundamental for autonomous robots in unstructured environments. Unlike fixed-base systems, mobile manipulators couple base mobility with arm dexterity. This additional base mobility spans the entire room and exponentially expands the search space, making valid kinematic solutions under strict articulation and collision constraints highly sparse~\cite{khatib1999mobile,mittal2022articulated,sleiman2023versatile}. Learning reliable policies for whole-body mobile manipulation therefore necessitates datasets orders of magnitude larger than those sufficient for stationary manipulation.

Yet acquiring such data at scale remains an open challenge. Teleoperation~\cite{fu2024mobile} yields high-fidelity demonstrations but is labor-intensive and fundamentally unscalable. Simulation-based \ac{rl}~\cite{fu2023deep,xia2021relmogen} automates data collection but suffers from prohibitively expensive exploration and persistent sim-to-real gaps. Planning-based methods~\cite{sleiman2023versatile} ensure physical validity, yet their CPU-based implementations are computationally prohibitive: the \ac{akr} framework~\cite{jiao2025integration}, which unifies base, arm, and object kinematics into a single representation~\cite{jiao2021efficient}, generates only 60 trajectories per hour~\cite{zhang2025m}. End-to-end learning methods~\cite{li2023behavior,zhang2025m} attempt to sidestep these issues by replacing exhaustive search with rapid neural inference, but remain bottlenecked by scarce coordinated whole-body demonstrations. Collectively, these constraints have forced prior datasets to compromise on scale, diversity, or kinematic fidelity~\cite{pari2022surprising,wu2023tidybot,cui2025gapartmanip}, leaving generalizable policy learning for whole-body mobile manipulation largely unsolved.

The \ac{akr} framework represents a principled foundation for this task, yet its potential has been curtailed by throughput limitations~\cite{zhang2025m}. This bottleneck has fragmented research into narrow-purpose datasets~\cite{ceola2023lhmanip,fu2024mobile}, underscoring the need for a framework that bridges precise kinematic modeling with the throughput of modern parallel computing at scale.

We address this with \dataset, a scalable framework that integrates \ac{akr} modeling with \acs{gpu}-accelerated motion planning~\cite{sundaralingam2023curobo}. By batching trajectory optimization and collision checking on the \ac{gpu}, \dataset generates physically valid whole-body trajectories at 5,000 episodes per \acs{gpu}-hour---80$\times$ faster than CPU-based baselines---enabling a dataset of over 500k trajectories across 330 realistic scenes, diverse robot morphologies, and articulated objects. The framework further supports complex, multi-step interactions such as grasp switching in confined spaces, ensuring high-fidelity data generation without costly human demonstrations.

Beyond dataset construction, we empirically validate the necessity of this scale. Experiments with downstream policies reveal that even a single articulated-object task requires tens of thousands of demonstrations for current \ac{sota} methods to reach 80\% success, confirming that data scarcity---not algorithmic limitations---has been the fundamental binding constraint in learning whole-body mobile manipulation.

In summary, our contributions are: (i) a \acs{gpu}-accelerated \ac{akr} planner that generates physically valid whole-body trajectories at 5,000 episodes per \acs{gpu}-hour (80$\times$ speedup), effectively resolving the acquisition bottleneck; (ii) a comprehensive dataset comprising over 500k trajectories across 330 scenes, covering diverse articulated objects and robot embodiments; and (iii) empirical evidence that \ac{sota} policies (\eg, DP3~\cite{ze20243d}) require tens of thousands of demonstrations to achieve high success rates, underscoring the imperative for the scale that \dataset provides. Together, these contributions establish \dataset as the first framework to bridge high-performance planning and large-scale learning for whole-body mobile manipulation.

\begin{table*}[t!]
    \centering
    \small
    \setlength{\tabcolsep}{3pt}
    \caption{\textbf{Comparison of \dataset with existing mobile manipulation datasets.} Existing datasets are constrained by their acquisition methods: teleoperation yields high-fidelity but small-scale data, while scripted policies lack whole-body coordination. \dataset overcomes these limitations through \acs{gpu}-accelerated automated planning, simultaneously achieving large scale, broad diversity, and high-fidelity joint-space trajectories---a combination no prior dataset provides. ``\textbf{Coord.}'': presence of whole-body base-arm coordination.}
    \label{tab:dataset_comparison}
    \begin{tabular}{@{} l l r c r l l @{}}
        \toprule
        \textbf{Dataset}                                       & \textbf{Robot}       & \textbf{\# Episodes} & \textbf{Coord.} & \textbf{\# Scenes} & \textbf{Action}     & \textbf{Method}          \\
        \midrule
        RT-1 Robot Action~\cite{brohan2023rt}                  & Google Robot         & 73,499           & Yes             & 10                 & End-effector pose            & VR teleoperation         \\
        NYU VINN~\cite{pari2022surprising}                     & Hello Stretch        & 435              & Yes             & 3                  & End-effector pose            & Kinesthetic teaching     \\
        BC-Z~\cite{jang2022bc}                                 & Google Robot         & 39,350           & Yes             & 2--3               & End-effector pose            & VR teleoperation         \\
        ETH Agent Affordances~\cite{schiavi2023learning}       & Franka               & 120              & No              & 50                 & End-effector pose            & Scripted policy          \\
        QUT Dexterous Manip.~\cite{ceola2023lhmanip}           & Franka               & 200              & No              & 1                  & End-effector pose            & VR teleoperation         \\
        CMU Stretch~\cite{bahl2023affordances,mendonca2023structured} & Hello Stretch & 135              & No              & 10                 & End-effector pose            & Scripted Policy          \\
        ConqHose~\cite{mitrano2024conq}                        & Spot                 & 139              & Yes             & 3                  & End-effector vel.            & Scripted policy          \\
        DobbE~\cite{shafiullah2023bringing}                    & Hello Stretch        & 5,208            & Yes             & 216                & End-effector pose            & Tool-based teleoperation       \\
        Mobile ALOHA~\cite{fu2024mobile}                       & Mobile ALOHA         & 276              & Yes             & 5                  & Joint position          & Leader-follower teleoperation  \\
        TidyBot~\cite{wu2023tidybot}                           & TidyBot              & 24               & No              & 104                & Other               & Scripted primitives      \\
        \midrule
        \textbf{Ours (\dataset)}                               & \textbf{Multi-Robot} & \textbf{500,000} & \textbf{Yes}    & \textbf{330}       & \textbf{Joint position} & \textbf{Automatic motion planning} \\
        \bottomrule
    \end{tabular}
\end{table*}

\section{Related Work}

\subsection{Motion Planning for Mobile Manipulation}

\paragraph{Model-based planning}
Classical methods for whole-body mobile manipulation often rely on task-specific controllers, such as impedance control for door opening~\cite{jain2010pulling,karayiannidis2016adaptive,stuede2019door}, or general base-arm optimization for cluttered scenes~\cite{berenson2008optimization,gochev2012planning,bodily2017motion}. While effective in controlled settings, these methods require extensive hand-tuning and struggle to generalize across diverse object types and environments. The \ac{akr} framework~\cite{jiao2021efficient,jiao2021consolidating,jiao2025integration} unifies the base, manipulator, and object into a single kinematic chain, enabling constraint-aware planning in a unified configuration space. However, current implementations rely on CPU-based solvers~\cite{zhang2025m}, which are computationally prohibitive for large-scale generation and are typically limited to fixed grasp poses, thereby restricting their utility for diverse dataset creation.

\paragraph{Learning-based planning}
End-to-end deep \ac{rl} has been applied to whole-body control in simulation~\cite{xia2021relmogen,fu2023deep}, yet it remains highly sample-inefficient and prone to overfitting specific environments~\cite{sun2022fully}. \ac{il} offers a more data-efficient alternative~\cite{fu2024mobile,jang2022bc} but is fundamentally constrained by the availability of high-quality demonstrations. Both paradigms thus share a common dependency: large-scale datasets capturing physically valid whole-body motions, whose absence remains an unresolved bottleneck.

\subsection{Data Collection for Mobile Manipulation}

\paragraph{Simulated embodied AI platforms}
Simulators such as Habitat 2.0~\cite{szot2021habitat}, AI2-THOR~\cite{kolve2017ai2}, OmniGibson~\cite{li2023behavior}, and RoboHive~\cite{kumar2023robohive} provide photorealistic environments but often prioritize visual fidelity over physical accuracy. Interactions in these platforms are frequently reduced to scripted primitives that bypass the complexities of base-arm and object kinematics. While benchmarks such as ManiSkill-HAB~\cite{shukla2025maniskill} enable policy learning for mobile manipulation, they often restrict training to a narrow set of objects or single-scene layouts, lacking the environmental diversity required for robust generalization.

\paragraph{Teleoperation}
Human-guided teleoperation captures realistic behaviors but suffers from severe scalability issues. Early systems recorded only end-effector trajectories~\cite{wu2019teleoperation,yang2023moma}, omitting the joint-space data essential for whole-body motion. Modern platforms such as Mobile ALOHA~\cite{fu2024mobile}, Behavior Robot Suite~\cite{jiang2025behavior}, and TeleMoMa~\cite{dass2024telemoma} capture full-body motion but are constrained by operator fatigue and hardware limitations, thereby limiting datasets to the order of thousands. Data augmentation techniques such as MoMaGen~\cite{li2025momagen} aim to scale data by extracting task information from demonstrations and regenerating trajectories, but rely on decoupled planners that generate base and arm motions \textit{separately}, failing to synthesize truly \textit{whole-body} behavior.

\paragraph{Existing mobile manipulation datasets}
The constraints of these acquisition methods have led to a scarcity of large-scale datasets, as summarized in \cref{tab:dataset_comparison}. Existing resources such as BC-Z~\cite{jang2022bc} contain up to 39,350 episodes but primarily use stationary bases or decoupled base--arm control, while Mobile ALOHA~\cite{fu2024mobile} provides high-quality whole-body data but is limited to 276 demonstrations on a single platform. Consequently, current datasets are generally narrow in task coverage, robot diversity, or physical validity~\cite{pari2022surprising,wu2023tidybot,cui2025gapartmanip}. \dataset addresses these gaps by leveraging \acs{gpu}-accelerated planning to automate the generation of over 500k constraint-compliant whole-body trajectories across 330 scenes and multiple robot embodiments.

\section{Preliminaries}

We outline the \acs{akr}-based planning formulation that underlies \dataset, covering the \ac{akr} construction procedure, the motion planning problem formulation, and the integration of task and physical constraints.

\subsection{The \texorpdfstring{\acl{akr}}{}}\label{sec:prelim:akr}

\begin{figure}[b!]
    \centering
    \small
    \includegraphics[width=\linewidth]{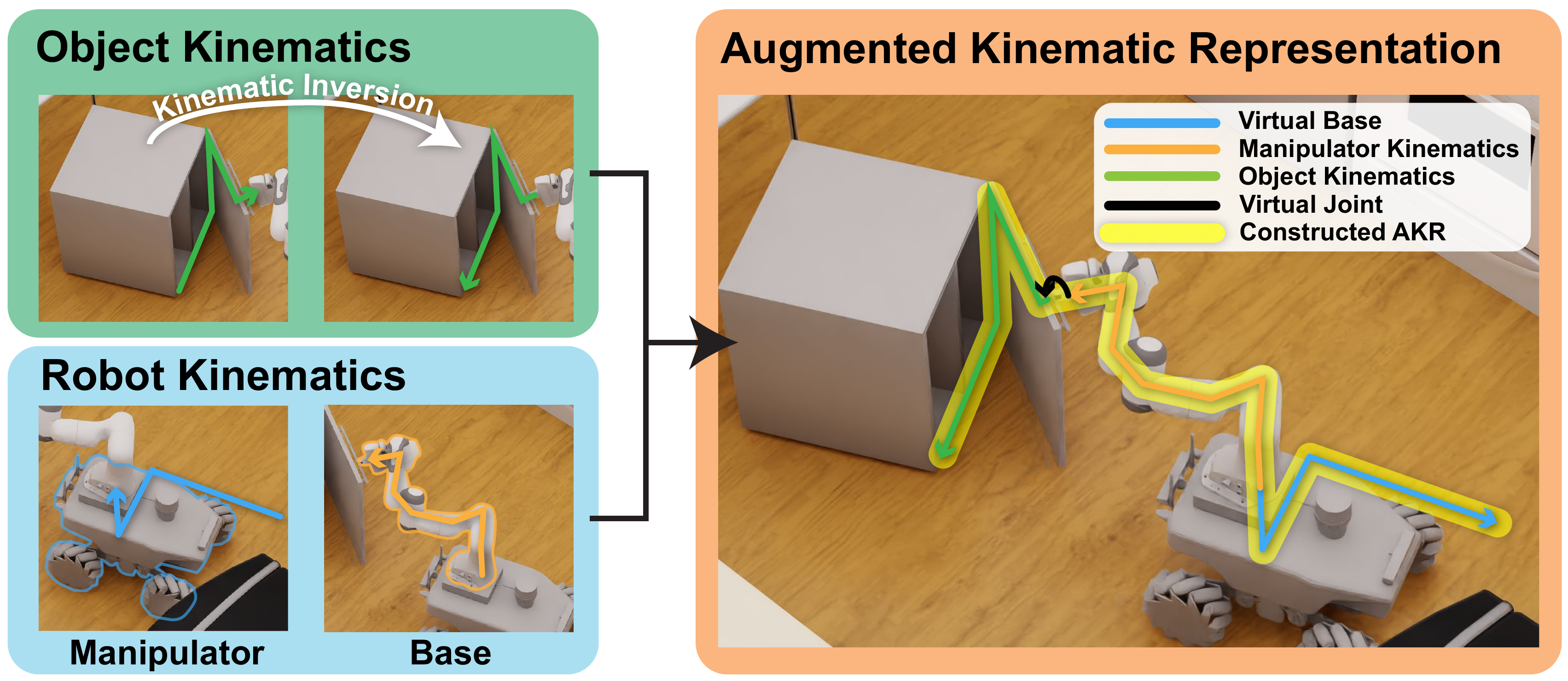}
    \caption{\textbf{An example of the \acs{akr} construction.} The \acs{akr} unifies independent kinematic trees into a single serial chain, enabling joint whole-body optimization of the base, arm, and object. The mobile base's planar motion is modeled via a \emph{virtual base} (blue), while a \emph{virtual joint} (black) couples the manipulator (orange) to the target object (green). For articulated objects, the kinematic tree is \emph{inverted} to reconfigure the kinematic root to the grasp point, forming a continuous chain (highlighted in yellow) rooted at the virtual world frame.}
    \label{fig:akr_construct}
\end{figure}

The \ac{akr} constructs a serial kinematic chain that integrates the mobile base, the manipulator, and the target object into a unified representation~\cite{jiao2025integration}, taking three inputs: (i) the robot's kinematic tree, (ii) the object's kinematic tree, and (iii) the transformation between the robot's end-effector and the object's attachable frame (\ie, the grasping pose).

As illustrated in \cref{fig:akr_construct}, the robot and object are initially represented as separate kinematic trees (\eg, via \ac{urdf}). To couple them into a single \ac{akr} chain, we attach the object to the robot by inserting a \emph{virtual joint} that encodes the grasp pose between the robot’s end-effector and the object; this requires inverting the object’s kinematic model so that the attachment link becomes the new kinematic root. Crucially, this inversion extends beyond reversing parent-child relationships: all associated transformations, including branching structures, must be rigorously updated, as revolute and prismatic joints typically define motion relative to the child link's frame. The geometry of these branching structures must likewise be preserved during trajectory optimization to ensure collision safety and physical feasibility. Implementation details for kinematic inversion and object-to-robot assembly are provided in \cref{supp:sec:akr_inversion_assembly}.

To enable joint optimization of mobile robot locomotion and manipulation, we introduce a \emph{virtual base} to model the mobile base's planar motion via two orthogonal prismatic joints and a revolute joint between the world frame and the robot's base, maintaining a strict serial kinematic structure throughout. \cref{fig:akr_construct} illustrates a constructed \ac{akr} for the door opening task. The resulting chain is rooted at the world link and terminates at the object's environmental anchor point (\eg, a fixed cabinet base). With the mobile base and manipulator embedded within this serial chain, their states---along with that of the object---are unified within the \ac{akr} configuration space, within which task goals and kinematic constraints are subsequently enforced during trajectory optimization.

\subsection{\texorpdfstring{\acs{akr}-Based Mobile Manipulation Planning}{AKR-Based Mobile Manipulation Planning}}

We formulate the whole-body mobile manipulation planning problem as finding a collision-free trajectory within the unified \ac{akr} configuration space that satisfies both kinematic and task-specific constraints. Formally, the \ac{akr} state is defined as:
\begin{align}
    \pmb{x} & = \bigl[\pmb{q}_B^{\mathsf T},\,\pmb{q}_M^{\mathsf T},\,\pmb{q}_O^{\mathsf T}\bigr]^{\mathsf T}
    \;\in\;\mathcal{X}_{\mathrm{free}},
    \label{eqn:akr_state}
\end{align}
where $\pmb{q}_B\in\mathbb{R}^3$ denotes the mobile base pose, $\pmb{q}_M\in\mathbb{R}^n$ represents the manipulator joint configuration ($n$ is the arm \ac{dof}), and $\pmb{q}_O\in\mathbb{R}^m$ is the joint state of the articulated object ($m$ is the object \ac{dof}, with $m=0$ for rigid objects). The planning objective is to generate a valid trajectory of length $T$: \(\pmb{x}_{1:T} = \langle \pmb{x}_{[1]}, \ldots, \pmb{x}_{[T]} \rangle \subset \mathcal{X}_{\mathrm{free}}\), where $\mathcal{X}_{\mathrm{free}}$ represents the collision-free configuration space. Following \citet{jiao2025integration}, we enforce the following constraints during trajectory optimization:
{%
\small%
\begin{align}
    h_{\mathrm{chain}}(\pmb{x}_{[t]})                                  & = 0,                             && \forall t=1,\dots,T,     \label{eqn:opt_chain} \\
    \|f_{\mathrm{task}}(\pmb{x}_{[T]}) - \pmb{g}_{\mathrm{goal}}\|_2^2 & \le \xi_{\mathrm{goal}},                                     \label{eqn:opt_goal}  \\
    \pmb{x}_{\min} \le \pmb{x}_{[t]}                                   & \le \pmb{x}_{\max},              && \forall t=1,\dots,T,     \label{eqn:cnt_jnt}   \\
    \|\Delta \pmb{x}_{[t]}\|_\infty                                    & \le \Delta \pmb{x}_{\max},       && \forall t=1,\dots,T-1,   \label{eqn:cnt_vel}   \\
    \|\Delta \dot{\pmb{x}}_{[t]}\|_\infty                              & \le \Delta \dot{\pmb{x}}_{\max}, && \forall t=2,\dots,T-1.   \label{eqn:cnt_acc}
\end{align}%
}%
\cref{eqn:opt_chain} enforces kinematic constraints imposed by the object's attachment to the environment, such as a revolute door's hinge connection or a sliding chair's planar constraint. \cref{eqn:opt_goal} ensures task completion by bounding the terminal state within a tolerance $\xi_{\mathrm{goal}}$ of the target goal $\pmb{g}_{\mathrm{goal}}$, as mapped by the task function $f_{\mathrm{task}}:\mathcal{X}\to\mathcal{G}$. \cref{eqn:cnt_jnt}--\cref{eqn:cnt_acc} impose physical limits on joint positions, velocities, and accelerations. Collision avoidance is handled implicitly via integrated self- and environment-collision checks within the underlying motion planner.

\begin{figure*}[t!]
    \centering
    \includegraphics[width=\linewidth]{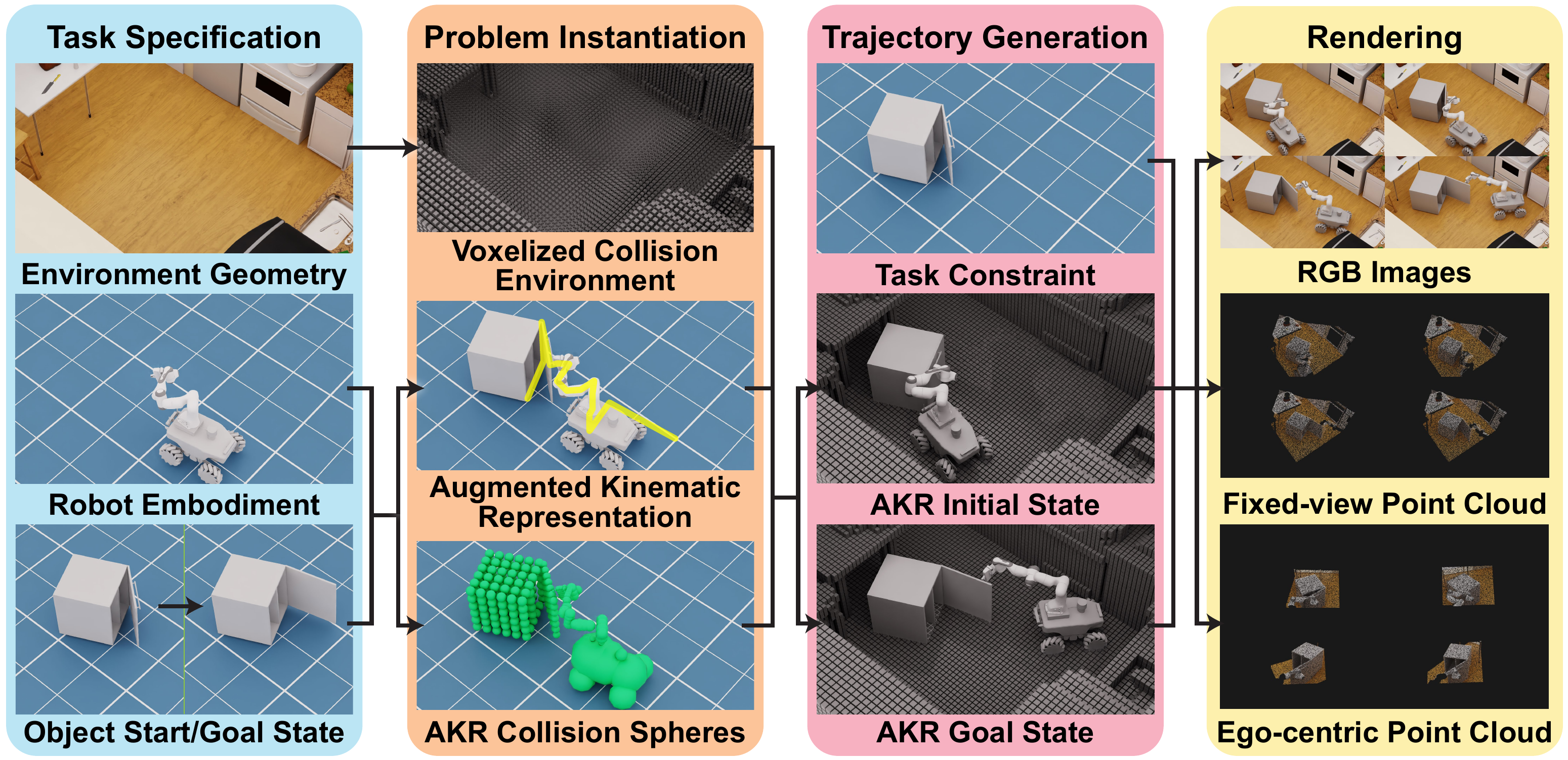}
    \caption{\textbf{The \dataset data generation pipeline.} Starting from a task specification triplet $(\mathcal{S}, \mathcal{O}, \mathcal{R})$ (left), \dataset proceeds through four stages: (i)~\textbf{Task Specification} defines the environmental, robotic, and object context; (ii)~\textbf{Problem Instantiation} transforms raw scene assets into planning-ready primitives via \acs{esdf} construction and \acs{akr} assembly with spherical collision approximations; (iii)~\textbf{Trajectory Generation} solves for optimal \acs{akr} states under task-specific constraints to produce physically valid whole-body motions; and (iv)~\textbf{Rendering} in NVIDIA Isaac Sim produces synchronized \acs{rgbd} sequences and point clouds. The resulting trajectories span diverse scenes, objects, and robot embodiments (right).}
    \label{fig:pipeline}
\end{figure*}

\section{The \dataset Data Generation Pipeline}

\dataset synthesizes large-scale, physically valid whole-body trajectory data through four integrated stages---task specification, problem instantiation, trajectory generation, and rendering---as illustrated in \cref{fig:pipeline}. We describe each stage in turn.

\subsection{Task Specification}

The pipeline accepts a task specification triplet $(\mathcal{S}, \mathcal{O}, \mathcal{R})$ that defines the semantic and geometric context of each mobile manipulation task.

\paragraph{Household scene layouts}
Each scene $\mathcal{S}$ encapsulates the geometric, visual, and semantic properties of structural elements such as walls, floors, and static appliances. All entities are anchored to a central world frame and associated with high-resolution visual and collision meshes. To maximize environmental diversity, our layouts are sourced via two complementary strategies: (i) procedural generation of interactive scenes with articulated appliances, and (ii) augmentation of existing scene datasets by substituting static appliances with functionally equivalent articulated counterparts. Details of scene construction are provided in \cref{supp:sec:scene_gen}.

\paragraph{Interactive objects}
The object set $\mathcal{O} = \mathcal{O}_{\text{rigid}} \cup \mathcal{O}_{\text{art}}$ includes both rigid and articulated entities. Rigid objects $o \in \mathcal{O}_{\text{rigid}}$ are characterized by watertight meshes and static grasp poses, whereas articulated objects $o \in \mathcal{O}_{\text{art}}$ require comprehensive \ac{urdf} descriptions specifying kinematic chains, joint limits, and inertial parameters. Crucially, the framework accounts for state-dependent grasp poses in articulated objects (\eg, handles moving during interaction); as detailed in \cref{sec:prelim:akr}, we re-root the object at the grasp point by inverting its kinematic tree, allowing the object to be attached to the end-effector and optimized jointly with the robot in a unified \ac{akr} configuration space.

\paragraph{Robot embodiments}
A robot embodiment $\mathcal{R}$ comprises a virtual mobile base and a manipulator, both defined via \ac{urdf}. To facilitate high-throughput \acs{gpu}-accelerated planning, each embodiment is augmented with a spherical approximation of its collision geometry, a self-collision mask to prune permanent contacts, and a joint-weight vector $\pmb{w} \in \mathbb{R}^{n+m+3}$ to modulate the optimization cost. The framework is embodiment-agnostic, as demonstrated across diverse platforms including a Franka arm on the Summit base, the R1 robot, and the TIAGo mobile manipulator.

\subsection{Problem Instantiation}

\dataset instantiates the planning problem by transforming raw scene geometries into structured representations that bridge environmental context with unified robot-object kinematics.

\paragraph{Environment collision models}
To accelerate environmental queries, each scene is converted into an \ac{esdf}. Efficiency is further enhanced by restricting planning queries to an axis-aligned bounding box defined by the object's start and goal states, thereby focusing computation on the local workspace and minimizing overhead.

\paragraph{\texorpdfstring{\ac{akr}}{} construction}
The \ac{akr} construction integrates the processed object model with the robot's kinematics into a single kinematic chain. To account for varying scene contexts, objects are first resized to align with the environment, with link components merged into single meshes for uniform scaling. Joint origins are then rigorously recalculated to compensate for the resulting spatial shifts and maintain kinematic integrity. This preprocessed model is coupled to the robot's end-effector via a virtual joint at the grasp pose, yielding a unified model $\mathcal{K}_{\text{akr}}$ in which the object effectively becomes a kinematic extension of the robot.

\paragraph{Collision processing}
For high-throughput \acs{gpu}-accelerated planning, link geometries are approximated using fitted spheres. Meshes are downscaled before sphere fitting to prevent volume overestimation and ensure conservative collision avoidance. When voxelization induces translational offsets, sphere cloud centroids are realigned with the original meshes to preserve geometric fidelity. Negligible collision pairs, such as permanently contacting adjacent links, are additionally masked to reduce computational overhead. To manage collision artifacts across task phases, we employ a dynamic strategy. In the \textit{approach} phase, environmental voxels intersecting the object are temporarily cleared and replaced with the high-resolution mesh, preventing discretization errors from blocking valid object grasp poses. In the \textit{manipulation} phase, the object transitions to a link of the \ac{akr} and its static environmental mesh is removed; only voxels lying strictly outside the object's current volume remain active, eliminating false-positive collisions with the object's initial state. \cref{supp:sec:sphere_fit} details the collision-sphere fitting and alignment procedure.

\subsection{Trajectory Generation}

\dataset synthesizes whole-body trajectories by formulating and solving a constrained optimization problem within the unified \ac{akr} configuration space, enabling precise goal specification while simultaneously enforcing task-specific constraints across the mobile base, manipulator, and object states.

\paragraph{Task objectives and goals}
The planning objective $\mathcal{J}$ minimizes both total travel distance and trajectory non-smoothness. Let $\pmb{x}_{1:T}$ denote the trajectory over $T$ time steps:
\begin{align}
    \mathcal{J}(\pmb{x}_{1:T}) &= \sum_{t=1}^{T-1} \big\|\pmb{w}_{v}\,\Delta{\pmb{x}}_{[t]} \big\|_2^2 + \sum_{t=2}^{T-1} \big\| \pmb{w}_{a}\,\Delta\dot{\pmb{x}}_{[t]} \big\|_2^2, \\
    \pmb{x}_{1:T}^\star &= \arg\min_{\pmb{x}_{1:T}} \mathcal{J}(\pmb{x}_{1:T}),
\end{align}
where diagonal weight matrices $\mathbf{W}_{v}$ and $\mathbf{W}_{a}$ modulate coordination strategies, such as prioritizing base stability during interaction. Task goals are defined according to object type: target $SE(3)$ poses for rigid object relocation, or specific joint configurations (\eg, door opening angles) for articulated objects.

\paragraph{Task constraints}
Trajectory constraints are derived from the semantic and physical relationships between the object and the scene. Rigid objects are modeled as free-floating, while heavy entities such as chairs are restricted to $SE(2)$ planar motion. For stationary articulated objects, a strict pose constraint is enforced on the \ac{akr} end-effector to penalize deviations from the object's base link, effectively modeling its physical attachment to the environment.

\paragraph{Optimization problem formulation}
Valid \ac{akr} start and goal configurations are computed by solving \ac{ik} for the respective object states. To manage computational overhead while ensuring diverse configuration-space coverage, similar \ac{ik} solutions are clustered in joint space, retaining a compact set of representative candidate configurations. For complex tasks where kinematic limits or collisions prevent a continuous grasp (\eg, opening a dishwasher in a confined space), \dataset employs a multi-stage strategy, sampling an intermediate state $\phi_{\mathrm{mid}}$ to connect two trajectory segments---$[\phi_0 \to \phi_{\mathrm{mid}}]$ and $[\phi_{\mathrm{mid}} \to \phi_T]$---via a collision-free re-grasp action. Qualitative examples are provided in \cref{supp:sec:grasp_switch}.

\paragraph{Trajectory post-processing}
Optimized trajectories are filtered to remove solutions that violate task constraints and ensure kinematic consistency. Each waypoint $\pmb{x}_{[t]}$ is validated against the required constraints. For stationary articulated objects, we evaluate translational deviation $d$ and rotational deviation $\theta$ for the object-world attachment:
\begin{equation}\begin{aligned}
    d &= \|p(\pmb{x}_{[t]}) - p(\pmb{x}_{\mathrm{ref}})\|_2,\\
    \theta &= \arccos(2\langle r(\pmb{x}_{[t]}), r(\pmb{x}_{\mathrm{ref}})\rangle^2 - 1),
\end{aligned}\end{equation}
where $p(\cdot)$ and $r(\cdot)$ denote the positional and rotational components of the \ac{akr} \ac{fk}. For planar constraints, vertical displacement $d_z$ and orientation deviation $\theta_{\mathrm{planar}}$ are additionally bounded:
\begin{equation}\begin{aligned}
    d_z &= |p_z(\pmb{x}_{[t]}) - p_z(\pmb{x}_{\mathrm{ref}})|,\\
    \theta_{\mathrm{planar}} &= \|\psi(\pmb{x}_{[t]}) - \psi(\pmb{x}_{\mathrm{ref}})\|_2,
\end{aligned}\end{equation}
where $p_z(\cdot)$ is the $z$-axis translation and $\psi(\cdot)$ represents roll and pitch angles. Trajectories violating these thresholds are discarded, ensuring the final dataset contains only stable, physically plausible whole-body motions.

\subsection{Rendering}

The final pipeline stage uses NVIDIA Isaac Sim to synthesize high-fidelity multi-modal observations from validated trajectories. Synchronized egocentric and fixed-viewpoint \ac{rgbd} cameras are configured on both the robot platform and within the environment to ensure multi-perspective coverage. At each waypoint $\pmb{x}_{[t]}$, RGB and depth images are rendered and projected into 3D point clouds within the simulation world coordinate frame, pairing every joint-space configuration with its corresponding geometric and visual context.

The rendering framework is designed for extensibility: camera placements are fully customizable, and saved trajectories can be replayed under varying lighting conditions, camera configurations, or sensor modalities. The resulting dataset provides a robust foundation for diverse downstream tasks, including \ac{il}~\cite{fang2019survey,hua2021learning}, visual servoing~\cite{sun2018review,janabi2010comparison}, and affordance detection~\cite{chu2019toward,do2018affordancenet}.

\section{Experiments}

We evaluate \dataset along two dimensions: (i) characterizing the diversity and generation efficiency of the synthesized dataset; and (ii) empirically investigating how data scale and diversity affect policy learning for whole-body mobile manipulation.

\subsection{Dataset Statistics and Diversity Analysis}

\begin{figure*}[t!]
    \centering
    \small
    \begin{subfigure}{\linewidth}
        \centering
        \includegraphics[width=\linewidth]{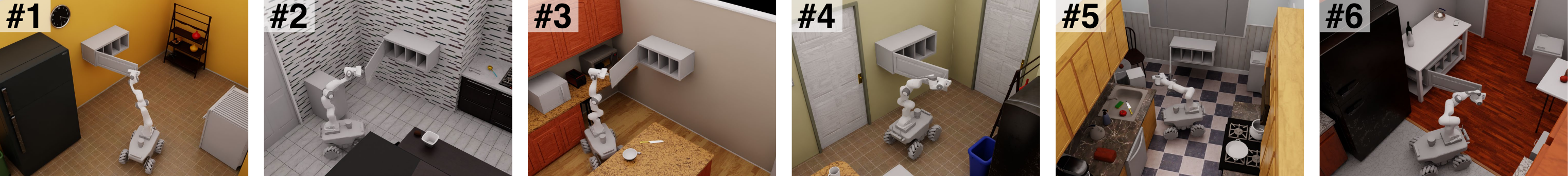}
        \caption{Scene layouts (\#1--\#6), ordered by increasing spatial confinement.}
        \label{fig:scenes}
    \end{subfigure}\\
    \begin{subfigure}{0.333\linewidth}
        \centering
        \includegraphics[width=\linewidth]{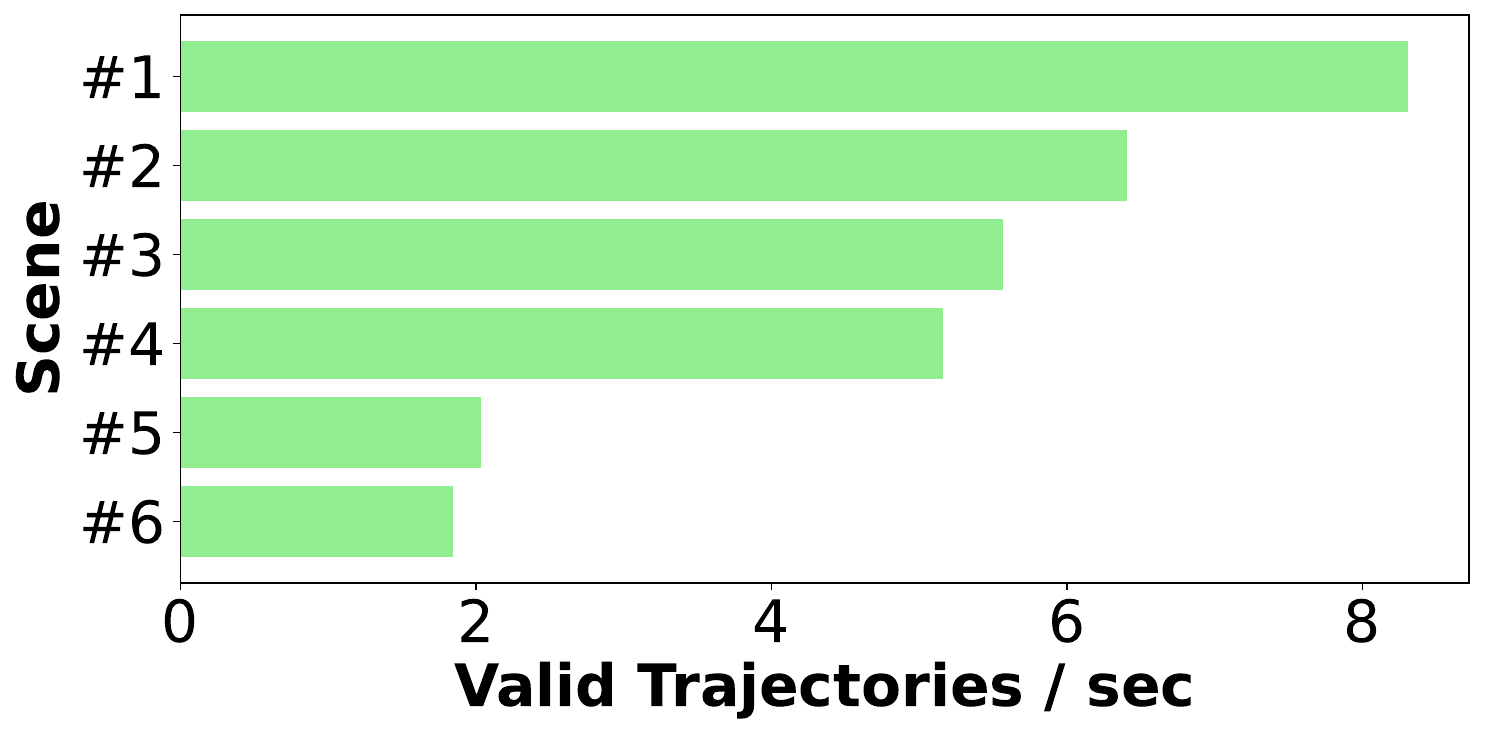}
        \caption{Generation throughput.}
        \label{fig:time}
    \end{subfigure}%
    \begin{subfigure}{0.333\linewidth}
        \centering
        \includegraphics[width=\linewidth]{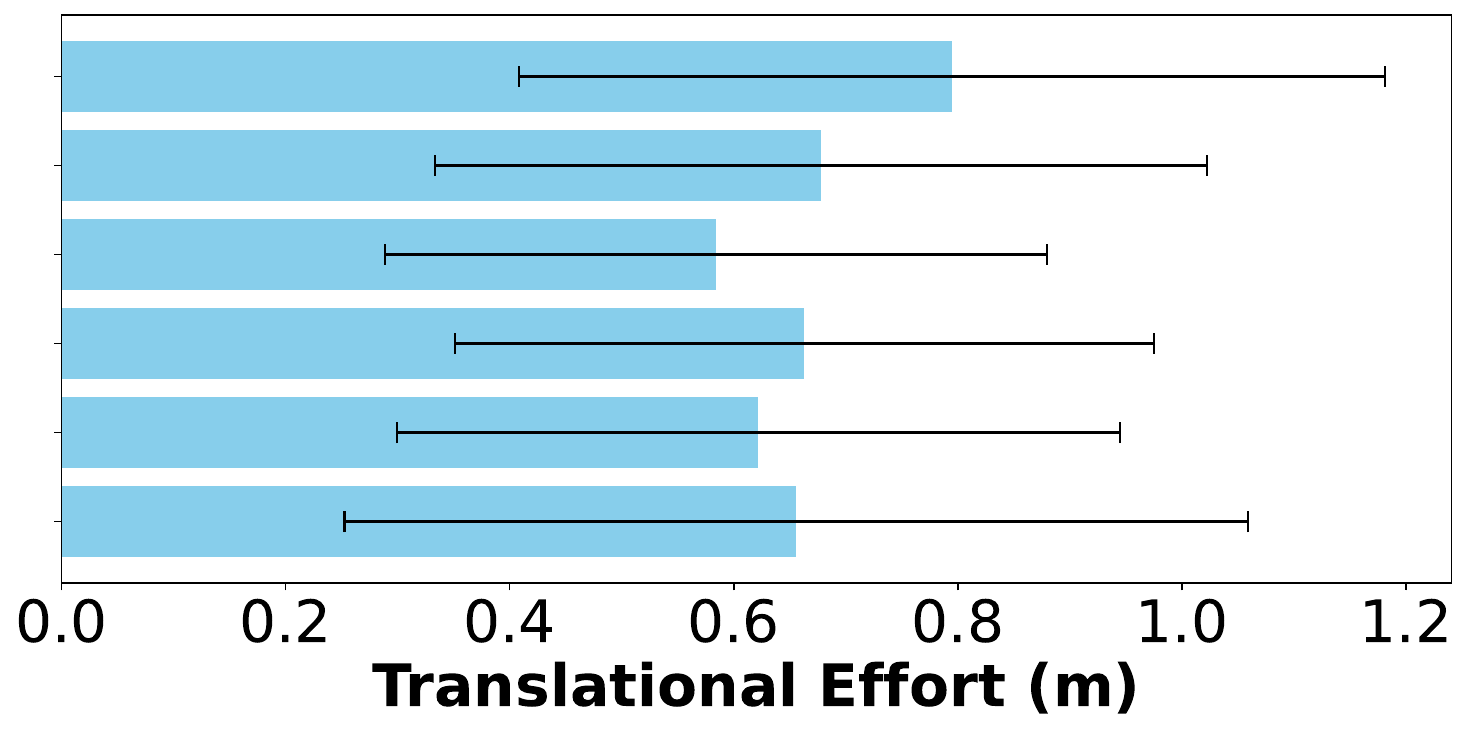}
        \caption{Ave. translational effort of the base.}
        \label{fig:trans_effort}
    \end{subfigure}%
    \begin{subfigure}{0.333\linewidth}
        \centering
        \includegraphics[width=\linewidth]{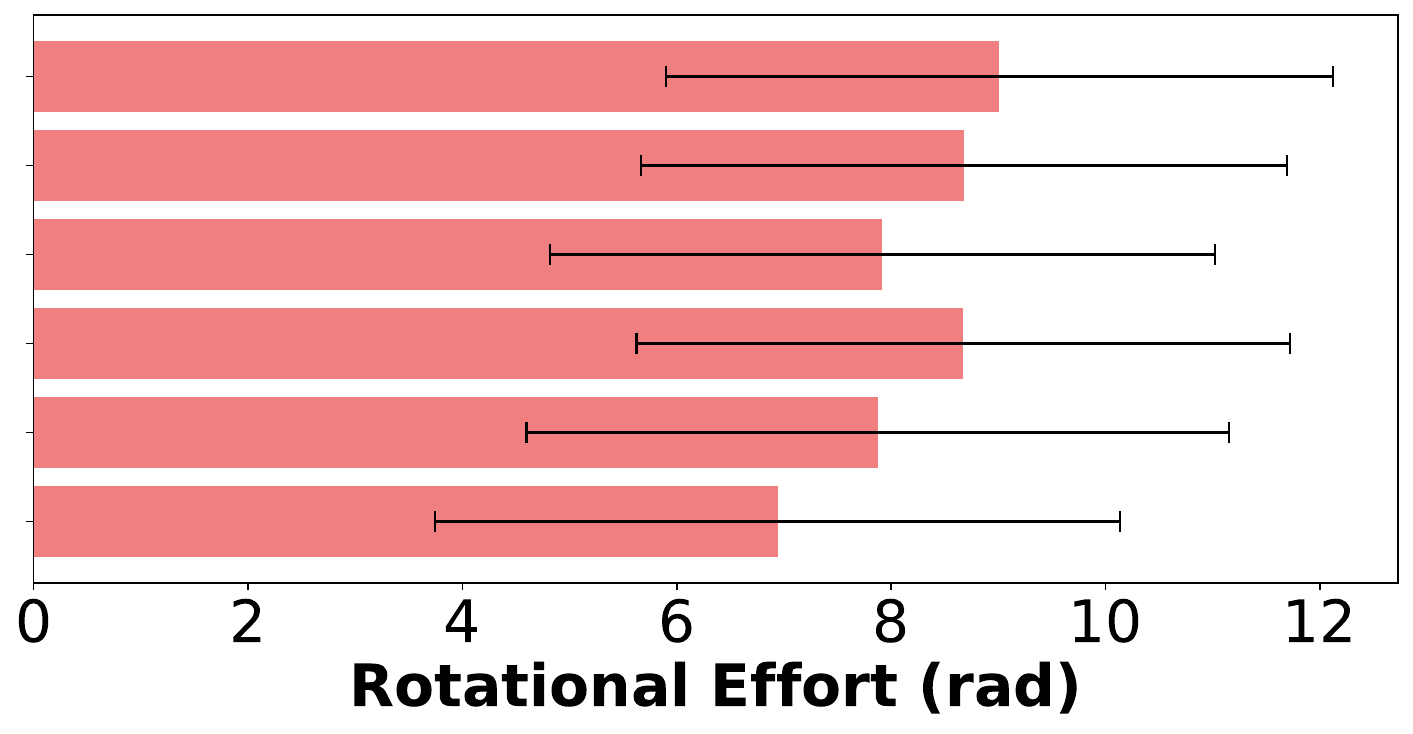}
        \caption{Ave. rotational effort of the manipulator.}
        \label{fig:rot_effort}
    \end{subfigure}%
    \caption{\textbf{Trajectory generation performance across six representative household scenes.} (a) Test scenes with increasing spatial confinement. (b) Generation throughput (valid trajectories per second) decreases as scene clutter increases collision-checking overhead. (c) Average translational effort of the mobile base per trajectory (error bars: standard deviation). (d) Average rotational effort of the manipulator, reflecting compensatory whole-body motion in constrained environments.}
    \label{fig:full}
\end{figure*}

\begin{figure}[t!]
    \centering
    \small
    \includegraphics[width=\linewidth,trim=0 0 0 40,clip]{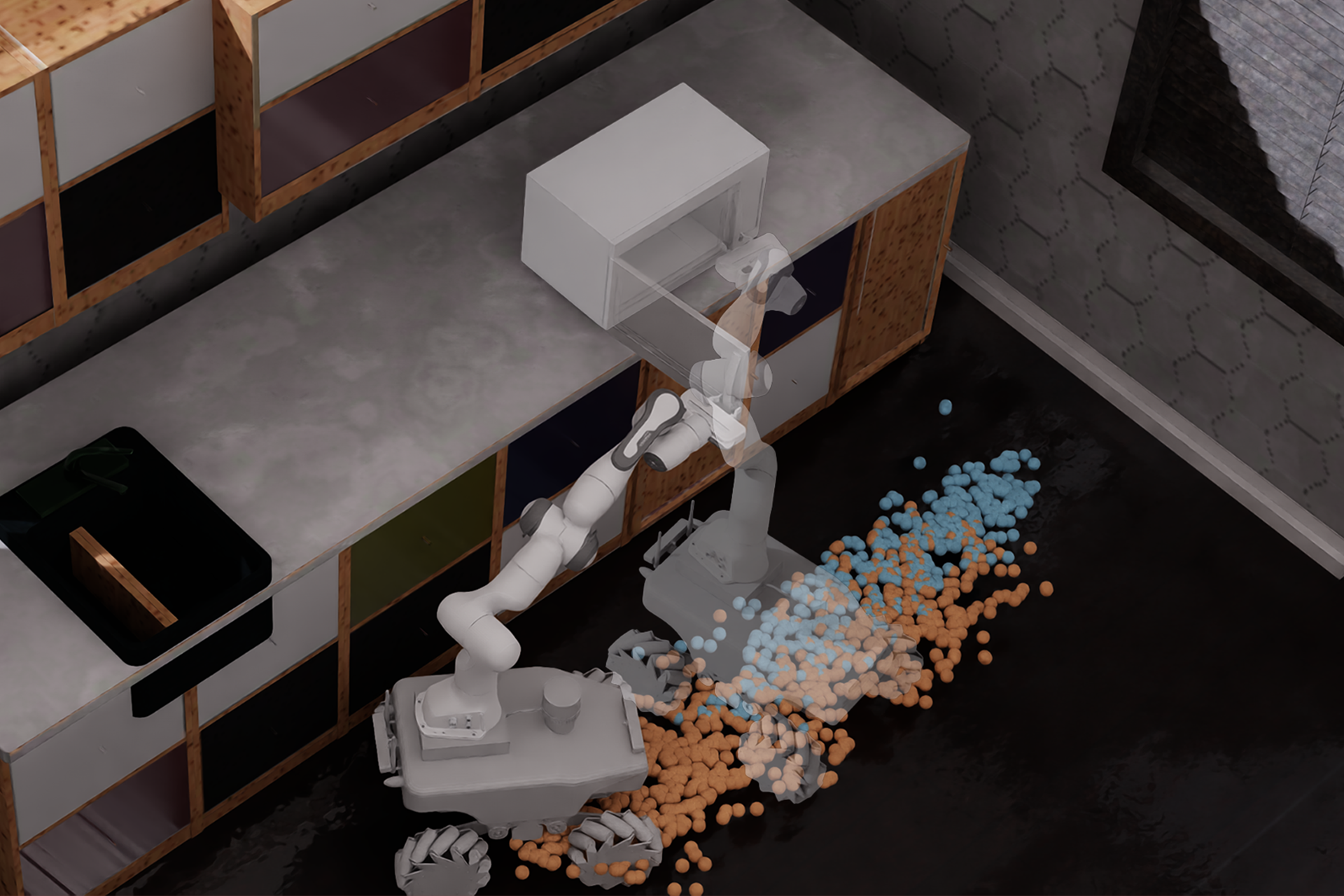}
    \caption{\textbf{Distribution of trajectory base positions.} Blue and orange spheres denote start and goal base placements, respectively, illustrating the broad spatial coverage achieved by the \acs{ik} clustering strategy.}
    \label{fig:ik_spread}
\end{figure}

\dataset integrates existing virtual household environments~\cite{kolve2017ai2,raistrick2024infinigen} populated with articulated objects sourced from PartNet-Mobility~\cite{xiang2020sapien}. Leveraging three distinct robot platforms---Summit Franka, TIAGo, and R1---we generated over 500k physically valid trajectories. Each trajectory comprises 30 joint-space waypoints accompanied by synchronized multi-modal observations, including \ac{rgbd} images and point clouds (4,096 points per frame) rendered at 120 frames per trajectory.

\paragraph{Grasp and configuration diversity}
To ensure broad coverage of the configuration space, each object is paired with $\sim$20 AO-Grasp~\cite{morlans2024grasp} annotations. We compute approximately 30 \ac{ik} solutions per grasp and cluster them in joint space to retain a diverse set of representative start states. This sampling strategy (\cref{fig:ik_spread}) ensures that trajectories span a broad distribution of feasible robot base placements.

\paragraph{Pipeline performance benchmarks}
We benchmark the trajectory generation pipeline across six representative household scenes (\cref{fig:scenes}) with varying spatial constraints. Less cluttered layouts yield higher throughput, whereas confined spaces increase collision-checking overhead and reduce feasible \ac{ik} counts (\cref{fig:time}). To further characterize planner behavior, we measure the average translational motion of the base (\cref{fig:trans_effort}) and cumulative arm rotation (\cref{fig:rot_effort}); these metrics reflect the planner's ability to synthesize compensatory whole-body motions in diverse and restrictive environments.

\subsection{Policy Learning Setup}

We evaluate the utility of the synthesized trajectories by training \ac{il} policies for whole-body coordination across three representative architectures.

\paragraph{Physical simulation}
All experiments are conducted in Isaac Sim, utilizing its \acs{gpu}-accelerated PhysX engine to simulate complex articulated interactions with high-fidelity physical feedback and synchronized sensor rendering.

\paragraph{Agent and observation space}
We employ the Summit Franka mobile manipulator as our primary agent. The policy architecture is DP3~\cite{ze20243d}, a \ac{sota} diffusion-based method used to benchmark data scaling laws. To demonstrate that the benefits of \dataset are model-agnostic, we additionally evaluate DP (RGB-based Diffusion Policy)~\cite{chi2025diffusion} and ACT (Transformer Policy)~\cite{fu2024mobile}. The observation space consists of fused point clouds (4,096 points) aggregated from egocentric and fixed \ac{rgbd} cameras, together with proprioceptive states (joint positions and base pose). All visual inputs are rendered at $320 \times 240$ resolution.

\paragraph{Training and evaluation}
Models are trained for 300 epochs with a batch size of 256 using the AdamW optimizer ($lr = 1 \times 10^{-4}$). Policies are evaluated on a microwave door opening task; a trial is successful if the door reaches the target angle within 300 steps, with success rates averaged over 50 randomized trials per setting. Complete hyperparameters are reported in \cref{supp:sec:automoma_additional_details,supp:sec:experimental_details}.

\begin{figure*}[t!]
    \centering
    \small
    \begin{subfigure}{0.33\linewidth}
        \centering
        \includegraphics[width=\linewidth]{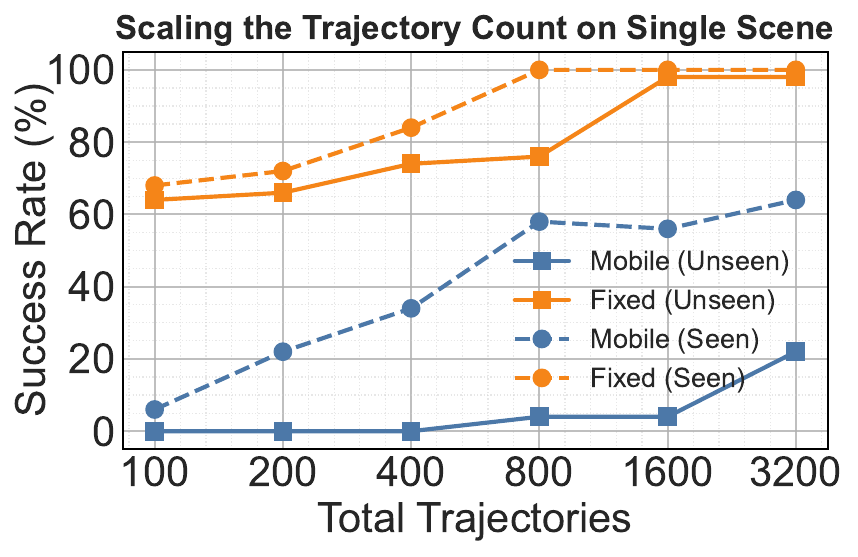}
        \caption{Fixed \vs Mobile Base on single scene.}
        \label{fig:traj_count_single}
    \end{subfigure}%
    \begin{subfigure}{0.33\linewidth}
        \centering
        \includegraphics[width=\linewidth]{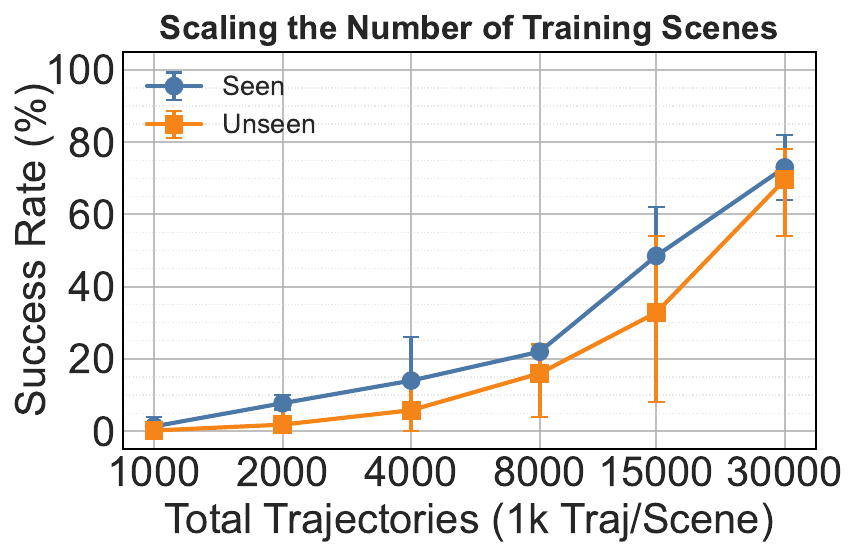}
        \caption{Scene count scaling with 1k traj/scene.}
        \label{fig:scene_count}
    \end{subfigure}%
    \begin{subfigure}{0.33\linewidth}
        \centering
        \includegraphics[width=\linewidth]{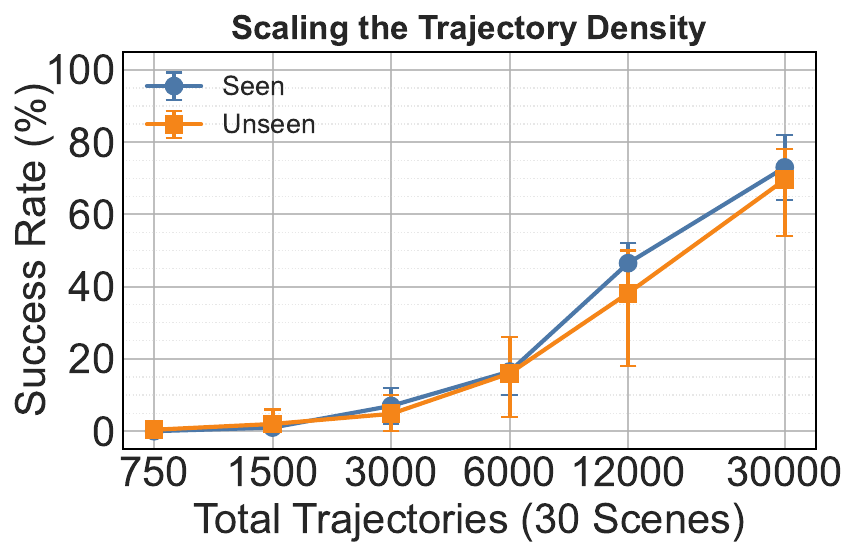}
        \caption{Trajectory density scaling on 30 scenes.}
        \label{fig:traj_count_multi}
    \end{subfigure}%
    \caption{\textbf{Data scaling experiments.} (a) In a single scene, the mobile base policy requires substantially more data than the fixed-base counterpart, with a persistent seen/unseen gap indicating manifold memorization. (b) Increasing scene diversity from 1 to 30 steadily improves generalization to unseen environments. (c) With 30 scenes, higher per-scene trajectory density further refines execution precision, enabling consistent generalization across seen and unseen scenes.}
    \label{fig:data_scaling}
\end{figure*}

\begin{figure}[t!]
    \centering
    \small
    \includegraphics[width=\linewidth]{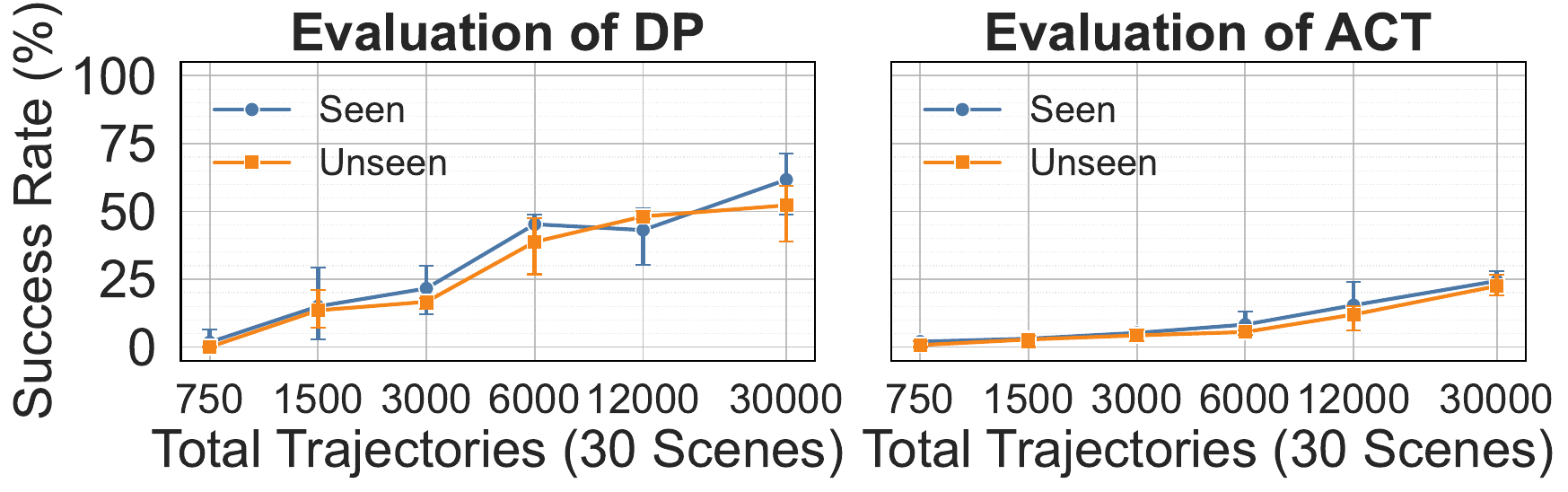}
    \caption{\textbf{Architectural generalization of \dataset.} When evaluated across the same 30-scene setup as DP3~\cite{ze20243d}, both DP~\cite{chi2025diffusion} and ACT~\cite{fu2024mobile} exhibit consistent performance gains with increasing trajectory density, demonstrating \dataset's compatibility with diverse whole-body \acs{il} architectures.}
    \label{fig:evaluation-dp-act}
\end{figure}

\begin{figure}[t!]
    \centering
    \small
    \includegraphics[width=\linewidth]{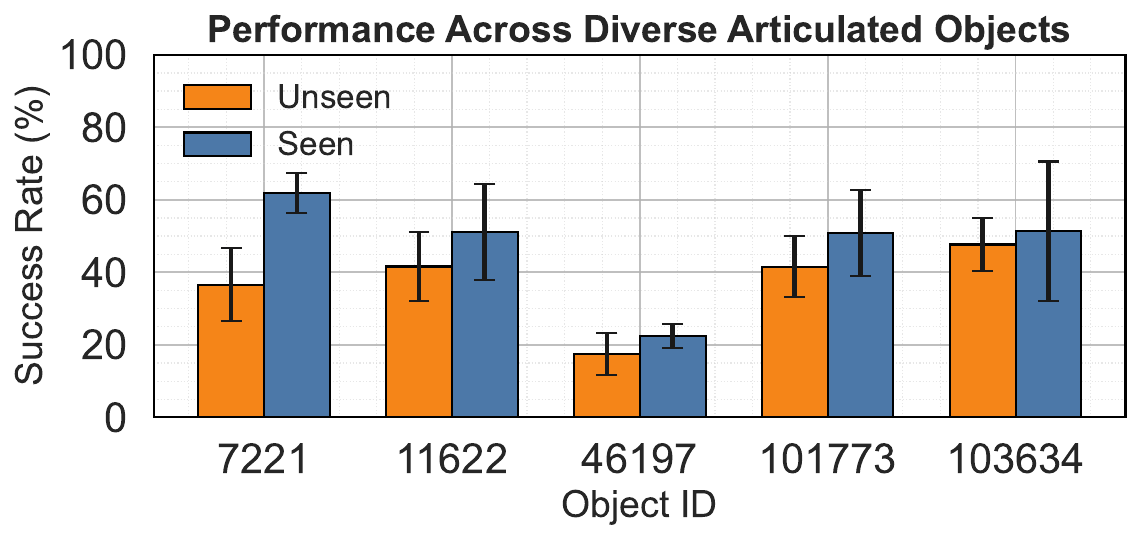}
    \caption{\textbf{Per-object success rates at 100k trajectories.} Success rates of the DP3 policy evaluated on five representative SAPIEN~\cite{xiang2020sapien} objects. The bar plot compares performance under unseen (orange) and seen (blue) environments.}
    \label{fig:per_object_success_100k}
\end{figure}

\subsection{Result Analysis}

We analyze policy performance across six dimensions to characterize how data scale and diversity affect whole-body mobile manipulation learning.

\paragraph{Configuration space complexity}
We first investigate the complexity gap between fixed-base and mobile manipulation (\cref{fig:traj_count_single}). The fixed-base robot achieves 100\% success with fewer than 800 trajectories, as its lower-dimensional configuration space is easily covered. In contrast, the mobile base policy saturates at approximately 70\% success on \emph{seen} configurations even with 3,200 trajectories. This performance gap stems from the complex 10-\ac{dof} base-arm coupling, exponentially expanding the search space and requiring massive data for robust coordination.

\paragraph{Local generalization}
Within a single-scene context, we compare performance on \emph{seen} and \emph{unseen} \ac{ik} states (\cref{fig:traj_count_single}). Despite strong performance on seen start configurations, the policy degrades significantly on novel \ac{ik} starts within the same workspace, indicating that high trajectory density in a single environment promotes manifold memorization rather than scene understanding. This underscores the need for broader diversity in the planning context.

\paragraph{Environmental diversity scaling}
Scaling from 1 to 30 scenes (\cref{fig:scene_count}) shows that policies trained on limited scenes fail to generalize to novel layouts due to overfitting to specific clutter and geometry. As scene diversity increases, success rates on \emph{unseen} environments improve steadily, demonstrating that geometric variety from the integrated iTHOR and Infinigen environments is the primary driver for learning transferable whole-body strategies.

\paragraph{Trajectory density in multi-scene scaling}
Fixing the environment count at 30 and varying per-scene trajectory density (\cref{fig:traj_count_multi}), we find that increasing trajectory density yields comparable improvements to expanding scene diversity. This higher density enables the model to capture a richer distribution of approach angles and motion variations, allowing the policy to generalize consistently across both seen and unseen scenes (${\sim}75\%$ success).

\paragraph{Architectural generalization}
We also evaluate \dataset on DP~\cite{chi2025diffusion} and ACT~\cite{fu2024mobile} (\cref{fig:evaluation-dp-act}). Both architectures exhibit consistent gains as trajectory density increases, though DP3 remains superior given its 3D modalities. This demonstrates the broad applicability of \dataset across different whole-body \ac{il} models. Further discussion is in \cref{supp:sec:scaling_100k}.

\paragraph{Performance stability across objects}
We evaluate five representative SAPIEN~\cite{xiang2020sapien} objects using the DP3 policy trained at the 100k trajectory scale (\cref{fig:per_object_success_100k}). The policy achieves over 50\% success on seen configurations for the majority of the evaluated objects (IDs 7221, 11622, 101773, and 103634), confirming robust scaling benefits across diverse kinematic constraints. Variance in specific objects (\eg, ID 46197) stems from articulation constraints limiting the robot workspace rather than data scarcity.

Qualitative inference rollouts, representative failure modes, and experiments on rigid-object picking are provided in \cref{supp:sec:inference_cases,supp:sec:pick_ablation}.

\section{Limitations and Conclusion}

\dataset provides a scalable framework that generates over 500k physically valid whole-body trajectories across diverse scenes, objects, and embodiments. While we validate these trajectories in the real world on a UR5-Ridgeback platform (see \cref{supp:sec:real_world}), several limitations remain. The current pipeline relies on known scene geometries and kinematics, and does not support dynamic human-robot interaction or deformable objects. Additionally, the sphere-based collision approximations required for \ac{gpu} acceleration can occasionally introduce geometric inaccuracies that cause execution failures (see \cref{supp:sec:failure_case}). Future work will integrate learning-based generation methods and develop community-driven tools that facilitate seamless extension to new robots and environments, further broadening \dataset's utility as a foundation for embodied AI.

\section*{Acknowledgment}
This work is supported in part by the National Science and Technology Innovation 2030 Major Program (2025ZD0219400), the National Natural Science Foundation of China (62376009 to Y.Z., and 52305007 to Z.J.), the PKU-BingJi Joint Laboratory for Artificial Intelligence, the Wuhan Major Scientific and Technological Special Program (2025060902020304), the Hubei Embodied Intelligence Foundation Model Research and Development Program, and the National Comprehensive Experimental Base for Governance of Intelligent Society, Wuhan East Lake High-Tech Development Zone.

{
    \small
    \bibliographystyle{ieeenat_fullname}
    \bibliography{reference_header,reference}
}

\clearpage
\appendix
\renewcommand\thefigure{A\arabic{figure}}
\setcounter{figure}{0}
\renewcommand\thetable{A\arabic{table}}
\setcounter{table}{0}
\renewcommand\theequation{A\arabic{equation}}
\setcounter{equation}{0}
\pagenumbering{arabic}%
\renewcommand*{\thepage}{A\arabic{page}}
\setcounter{footnote}{0}

\section{Additional Method and Implementation Details}

This section provides additional implementation details of the \ac{akr}-based planning pipeline, covering the \ac{akr} inversion and assembly procedure and the collision-sphere fitting process that supports efficient \acs{gpu}-accelerated collision checking.

\subsection{Detailed Implementation for AKR Inversion and Assembly}\label{supp:sec:akr_inversion_assembly}

The \ac{akr} inversion and assembly workflow comprises four steps: (a) \ac{urdf} preprocessing and kinematic inversion, (b) collision sphere generation, (c) object-to-robot attachment, and (d) selective self-collision checking. The process begins by applying a uniform scaling factor to the object \ac{urdf} to ensure physical consistency---particularly important when working with grasp datasets derived from point cloud or \ac{rgbd} data, where object models are often normalized to fit within a standardized bounding volume. Since each link may contain multiple geometric components, these are first merged into a single mesh and scaled accordingly. As mesh scaling alters the spatial relationships in the original kinematic chain $\mathcal{K}_{\text{raw}}$, all joint origins are subsequently recalculated to preserve valid relative transformations, yielding $\mathcal{K}_{\text{scaled}}$.

The tip link $\ell_{\text{tip}}$ corresponding to the grasping point is identified along with its parent joint. To attach the object as an extension of the robot, the kinematic structure is inverted by reassigning $\ell_{\text{tip}}$ as the new base link; the joint hierarchy from the original base $\ell_{\text{base}}$ to the tip is reversed while the rest of the tree is preserved, yielding $\mathcal{K}_{\text{inv}}$.

The attachment transformation between the robot and the object is computed from the grasp pose and the object's \ac{fk}. Let $T^{\text{base}}_{\text{tip}}$ denote the pose of $\ell_{\text{tip}}$ relative to $\ell_{\text{base}}$ under the joint configuration $\boldsymbol{q}_{\text{init}}$ corresponding to the grasp pose:
\begin{equation}
    T^{\text{base}}_{\text{tip}} = \text{FK}_{\mathcal{K}_{\text{scaled}}}(\boldsymbol{q}_{\text{init}}, \ell_{\text{tip}}).
\end{equation}
Given the grasp pose $T^{\text{base}}_{\text{tcp}}$ specifying the robot's \ac{tcp} frame relative to the object's base link, the final attachment transformation is:
\begin{equation}
    T^{\text{tcp}}_{\text{tip}} = \left( T^{\text{base}}_{\text{tcp}} \right)^{-1} \cdot T^{\text{base}}_{\text{tip}}.
\end{equation}
This transformation is applied as a fixed joint between the robot's \ac{tcp} and the object's new base link, yielding a unified kinematic model in which the robot and object form a single tree structure.

Finally, the additional self-collision link pairs introduced by the attached object are identified selectively, avoiding full recomputation of the entire self-collision matrix and thereby reducing computational overhead while maintaining sufficient coverage for motion planning. Detailed pseudocode is provided in \cref{supp:alg:akr_construction}.

\begin{algorithm}[ht]
    \caption{AKR Construction Procedure}
    \label{supp:alg:akr_construction}
    \SetAlgoLined
    \SetKwFunction{FMain}{akr\_construction}
    \SetKwProg{Fn}{Function}{:}{}
    \Fn{\FMain{robot, object, init\_state, scaling\_factor, sphere\_params, grasp\_pose, grasp\_link, sample\_n}}{
        \Call{update\_object\_state}{object, init\_state}\;
        \ForEach{link $\in$ object.links}{
            merged\_mesh $\gets$ \Call{merge}{link.geometries}\;
            scaled\_mesh $\gets$ \Call{scale}{merged\_mesh, scaling\_factor}\;
            link.visual $\gets$ scaled\_mesh\;
            link.collision $\gets$ \Call{sphere\_fit}{scaled\_mesh, sphere\_params}\;
        }
        \ForEach{joint $\in$ object.joints}{
            tf $\gets$ \Call{get\_tf}{object, joint.child, joint.parent}\;
            joint.origin $\gets$ \Call{update}{tf, scaling\_factor}\;
        }
        scaled\_object $\gets$ object\;
        fk\_pose $\gets$ \Call{fk}{scaled\_object, grasp\_link}\;
        attached\_origin $\gets$ grasp\_pose$^{-1}$ $\cdot$ fk\_pose\;
        inversed\_object $\gets$ \Call{inverse}{scaled\_object, grasp\_link}\;
        akr $\gets$ \Call{attach}{robot, inversed\_object, attached\_origin}\;
        added\_link\_pairs $\gets$ \Call{filter}{akr.link\_pairs, robot.link\_pairs}\;
        sampled\_cfg $\gets$ \Call{sample\_cfg}{akr, sample\_n}\;
        added\_collision\_pairs $\gets$ \Call{check\_collision}{added\_link\_pairs, sampled\_cfg}\;
        akr.collision\_pairs $\gets$ \Call{union}{robot.collision\_pairs, added\_collision\_pairs}\;
    }
\end{algorithm}

\subsection{Collision-Sphere Fitting Procedure}\label{supp:sec:sphere_fit}

As the manipulated object becomes part of the robot representation via \ac{akr}, its collision geometry is approximated using a set of spheres, aligning with cuRobo's sphere-based representation and enabling \acs{gpu}-accelerated parallel collision checking. The fitting procedure proceeds as follows.
\begin{enumerate}
    \item \textbf{Mesh preprocessing:} Multiple geometric components within each link are merged into a single mesh.
    \item \textbf{Scaling and voxelization:} The merged mesh is uniformly scaled down slightly to ensure conservative collision checking, then voxelized into discrete occupied volumetric regions forming an occupancy grid.
    \item \textbf{Sphere fitting:} For each occupied voxel region, a sphere is positioned at the voxel centroid with diameter equal to the voxel edge length.
    \item \textbf{Spatial alignment:} When voxelization introduces translational offsets, the centroid of the fitted sphere cloud is realigned with that of the original mesh to mitigate discretization drift.
\end{enumerate}

The resulting sphere-based representation ensures computationally efficient collision queries throughout trajectory optimization.

\section{Interactive Scene Construction and Data Generation Details}

\subsection{Generation of Interactive Scenes}\label{supp:sec:scene_gen}

\begin{figure*}[t!]
    \centering
    \small
    \begin{subfigure}[b]{\linewidth}
        \centering
        \includegraphics[width=\linewidth]{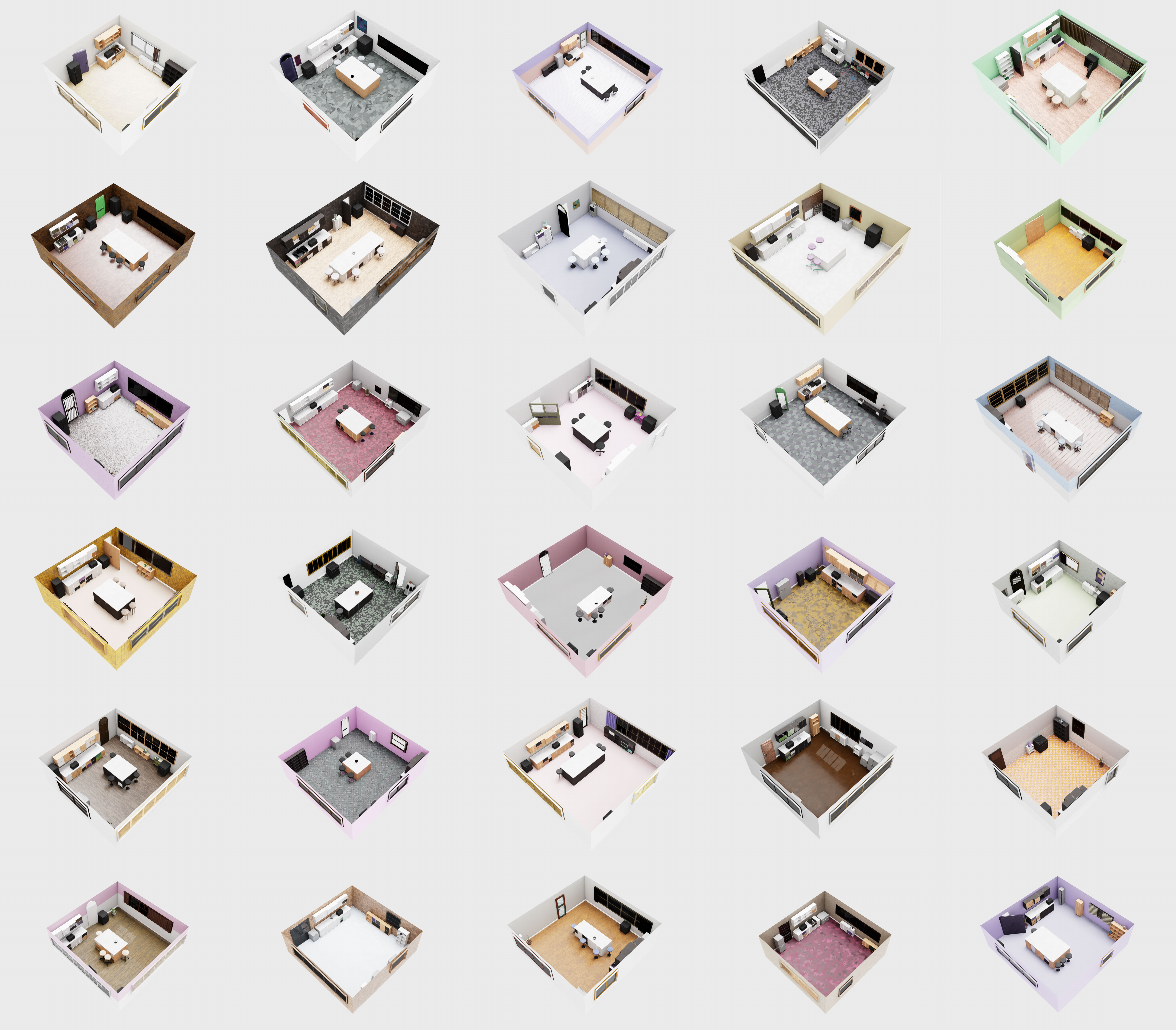}
        \caption{Training scenes (30 scenes) used for downstream policy training.}
        \label{fig:supp-train-scenes}
    \end{subfigure}%
    \\%
    \begin{subfigure}[b]{\linewidth}
        \centering
        \includegraphics[width=\linewidth]{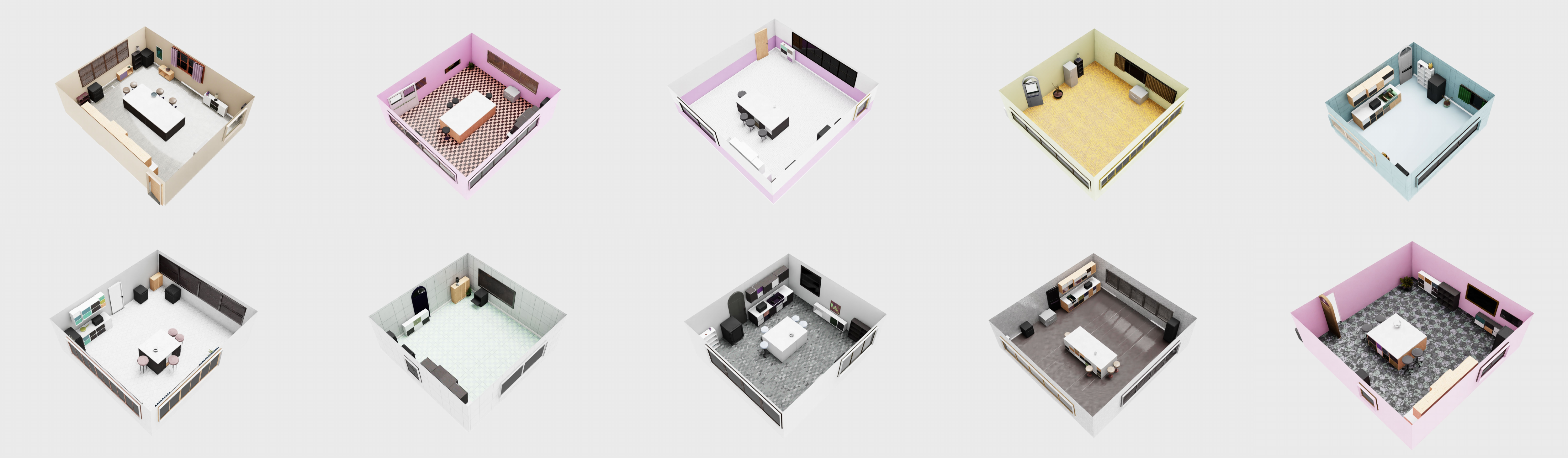}
        \caption{Test scenes (10 scenes) used for downstream evaluation.}
        \label{fig:supp-test-scenes}
    \end{subfigure}%
    \caption{\textbf{Downstream scene split.} 30 training and 10 test scenes are sampled from the 300 procedurally generated scenes for policy training and evaluation.}
    \label{fig:supp-scene-train-test}
\end{figure*}

\dataset leverages two complementary sources of interactive household environments: manually curated scenes and procedurally generated layouts.

\begin{figure*}[t!]
    \centering
    \small
    \begin{subfigure}[b]{\linewidth}
        \centering
        \includegraphics[width=\linewidth]{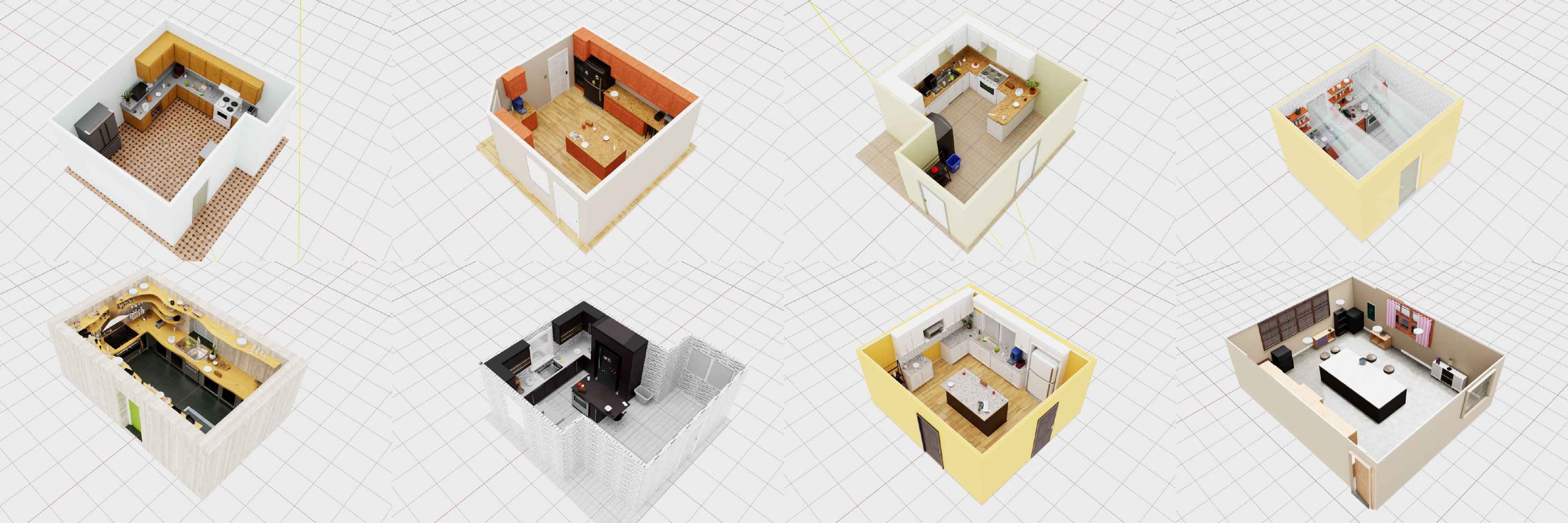}
        \caption{AI2-THOR (iTHOR) scene with SAPIEN asset replacement.}
        \label{fig:scene-ithor}
    \end{subfigure}%
    \\%
    \begin{subfigure}[b]{\linewidth}
        \centering
        \includegraphics[width=\linewidth]{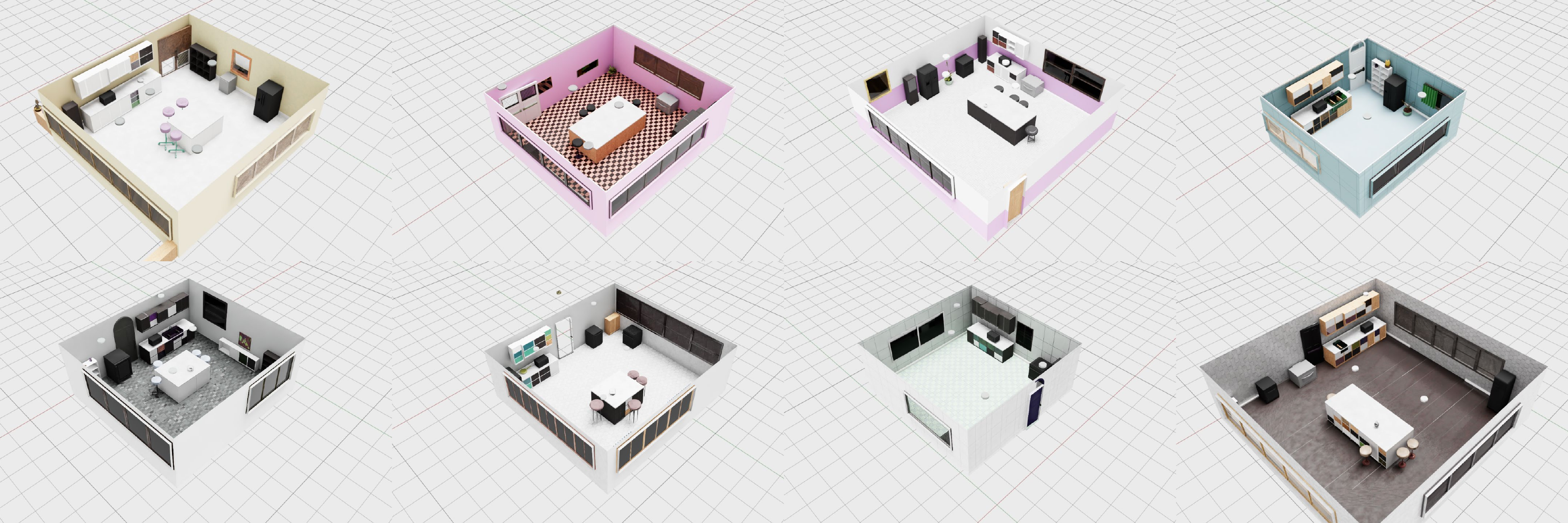}
        \caption{Procedurally generated scene from Infinigen.}
        \label{fig:supp:scene-infinigen}
    \end{subfigure}%
    \caption{\textbf{Two scene sources in \dataset.} (a) AI2-THOR scenes with articulated objects substituted from SAPIEN. (b) Procedurally generated layouts from Infinigen.}
    \label{fig:scene-sources}
\end{figure*}

The first source consists of 30 high-fidelity scenes derived from AI2-THOR~\cite{kolve2017ai2} (shown in \cref{fig:scene-ithor}), in which static objects such as microwave ovens, dishwashers, and cabinets are manually replaced with functionally equivalent articulated counterparts from the SAPIEN dataset~\cite{xiang2020sapien}. Replacements are carefully positioned to respect semantic context and physical plausibility, yielding coherent scenes suitable for targeted evaluation.

The second source comprises 300 diverse scenes generated using a custom Infinigen-based pipeline~\cite{raistrick2024infinigen}. Articulated object models are converted into static placeholder assets and imported into Infinigen; the generator supports controllable parameters---including object selection, placement optimization, and layout sparsity---to produce manipulation-friendly configurations. Generated layouts are exported in USD format for compatibility with \acs{gpu}-accelerated planning in cuRobo~\cite{sundaralingam2023curobo}, after which placeholders are replaced with the original articulated objects to restore full kinematic fidelity (\cref{fig:supp:scene-infinigen}).

Together, these 30 curated and 300 procedural scenes provide a rich and diverse foundation for training and evaluating whole-body mobile manipulation policies across a wide range of household contexts.

\subsection{Additional Details of Data Generation Pipeline}\label{supp:sec:automoma_additional_details}

\cref{supp:tab:automoma_config} details the configuration of the \dataset trajectory generation pipeline. Structured voxel or mesh collision fields are constructed with a safety margin of 0.3\,m and a voxelized distance resolution of 2\,cm. Both \ac{ik} and trajectory meshes are sampled at 5\,cm pitch to balance precision and scalability across hundreds of thousands of trajectories.

During \ac{ik} sampling and filtering, up to 10 retries are allowed per attempt, and a hybrid clustering strategy combining $k$-means and \acf{ap} is applied to select diverse candidate configurations. Grasp proposals (20 per object) and open-angle sampling (5--10 degrees) determine the target end-effector poses for trajectory goals, with position and rotation tolerances of 1\,cm ensuring consistency across generated sequences.

\section{Training Setup and Experimental Details}\label{supp:sec:experimental_details}

This section summarizes the complete experimental configuration for downstream policy learning. We evaluate three policy baselines---3D Diffusion Policy (DP3), image-based Diffusion Policy (DP), and Action Chunking with Transformers (ACT)---describing for each the observation and action structure, model architecture, and optimization setup.

\subsection{3D Diffusion Policy (DP3)}

\paragraph{Observation, action, and temporal structure}
DP3~\cite{ze20243d} represents the scene with a downsampled point cloud of 4,096 points (XYZ coordinates) together with an 11-dimensional proprioceptive state capturing mobile base and arm joint configurations. A temporal history of three consecutive observation steps is concatenated before being passed to the encoder, providing short-horizon context important for resolving the kinematic coupling between base placement and arm motion. The model predicts future actions over a horizon of eight steps, of which six are executed per re-planning cycle, with an 11-dimensional action vector covering the full whole-body \ac{dof}.

\paragraph{Policy architecture}
A PointNet-based encoder with optional spatial cropping (crop shape $[80\times80]$) accepts 3-channel input and produces a 64-dimensional latent with LayerNorm applied both within and at the output; color and normal channels are disabled to ensure simulation-to-real consistency. The latent conditions a U-Net diffusion backbone via Feature-wise Linear Modulation (FiLM) units applied at the downsampling, bottleneck, and upsampling stages, with channel dimensions $[512, 1024, 2048]$, kernel size 5, group normalization with 8 groups, and a 64-dimensional diffusion-step embedding.

\paragraph{Diffusion scheduler}
Training uses 100 diffusion steps with a DDIMScheduler, squared-cosine beta schedule, and sample-prediction parameterization. Inference requires only 10 denoising steps, providing a favorable accuracy--speed trade-off for real-time motion generation.

\paragraph{Optimization and training}
All hyperparameters are listed in \cref{tab:dp3_hyperparameter}. AdamW is used with $\text{lr}=1\times10^{-4}$, betas $(0.95, 0.999)$, weight decay $1\times10^{-6}$, a cosine-decay schedule with 500 warmup steps, and a batch size of 512. Exponential Moving Average (EMA) with power 0.75 and maximum value 0.9999 stabilizes training. All experiments use 300 epochs with a fixed random seed of 42.

\subsection{Image-based Diffusion Policy (DP)}

\paragraph{Observation, action, and temporal structure}
DP~\cite{chi2025diffusion} replaces the point cloud with raw RGB observations from three camera views (ego top-down, ego wrist, and fixed), retaining the same 11-dimensional proprioceptive state, three-step temporal history, 8-step prediction horizon, 6 executed steps per cycle, and 11-dimensional action vector as DP3.

\paragraph{Policy architecture}
Each camera stream is independently encoded by a ResNet-18 backbone with group normalization, standard ImageNet normalization, and random crop augmentation; weights are not shared across views and pretrained weights are not used. Concatenated multi-view features condition a U-Net diffusion backbone with channel dimensions $[256, 512, 1024]$, kernel size 5, 8 normalization groups, a 128-dimensional diffusion-step embedding, and conditional prediction scaling.

\paragraph{Diffusion scheduler}
DP uses a DDPMScheduler with a squared-cosine beta schedule, $\beta_{\text{start}}=0.0001$, $\beta_{\text{end}}=0.02$, epsilon-prediction parameterization, and clipped samples. All 100 denoising steps are performed at inference, which is feasible since the image-based variant carries no point cloud preprocessing overhead.

\paragraph{Optimization and training}
All hyperparameters are listed in \cref{tab:dp_hyperparameter}. AdamW is used with $\text{lr}=1\times10^{-4}$, betas $(0.95, 0.999)$, weight decay $1\times10^{-6}$, a cosine-decay schedule with 500 warmup steps, and a batch size of 128. Half-precision training (\texttt{fp16}) is enabled to reduce memory consumption. EMA (power 0.75, max 0.9999) is applied identically to the DP3 setup, for 300 epochs.

\subsection{Action Chunking with Transformers (ACT)}

\paragraph{Observation, action, and temporal structure}
ACT~\cite{fu2024mobile} operates on the same three camera views as DP. Rather than maintaining a temporal observation history, ACT encodes a single observation and predicts a \emph{chunk} of 6 consecutive actions in one forward pass. Temporal aggregation across overlapping chunks is disabled; each chunk is executed open-loop before the next prediction. Mobile base commands are in a relative coordinate frame, and the 11-dimensional action vector covers the full whole-body \ac{dof}.

\paragraph{Policy architecture}
Each image is encoded by a ResNet-18 backbone with sinusoidal position embeddings, feeding visual tokens into a CVAE-based Transformer encoder--decoder. The encoder has 4 layers with 8 attention heads and a 512-dimensional hidden space; the decoder has 7 layers with the same configuration and a 3,200-dimensional feedforward width. Dropout of 0.1 is applied throughout, with pre-norm, dilation, and mask-prediction disabled. A KL-divergence weight of 1.0 regularizes the CVAE latent.

\paragraph{Optimization and training}
All hyperparameters are listed in \cref{tab:act_hyperparameter}. The backbone and Transformer are jointly optimized with AdamW at $\text{lr}=1\times10^{-5}$ and global weight decay $1\times10^{-4}$. Training runs for 2,000 epochs on 100 demonstrations per task configuration with episode length 1,000 steps. No learning rate scheduler or EMA is applied; the longer training duration and lower learning rate suffice for stable convergence.

\section{Qualitative Results, Failure Analysis, and Real-World Validation}

\subsection{Representative Success and Failure Cases of Trajectory Generation}\label{supp:sec:failure_case}

\begin{figure*}[t!]
    \centering
    \small
    \includegraphics[width=\linewidth]{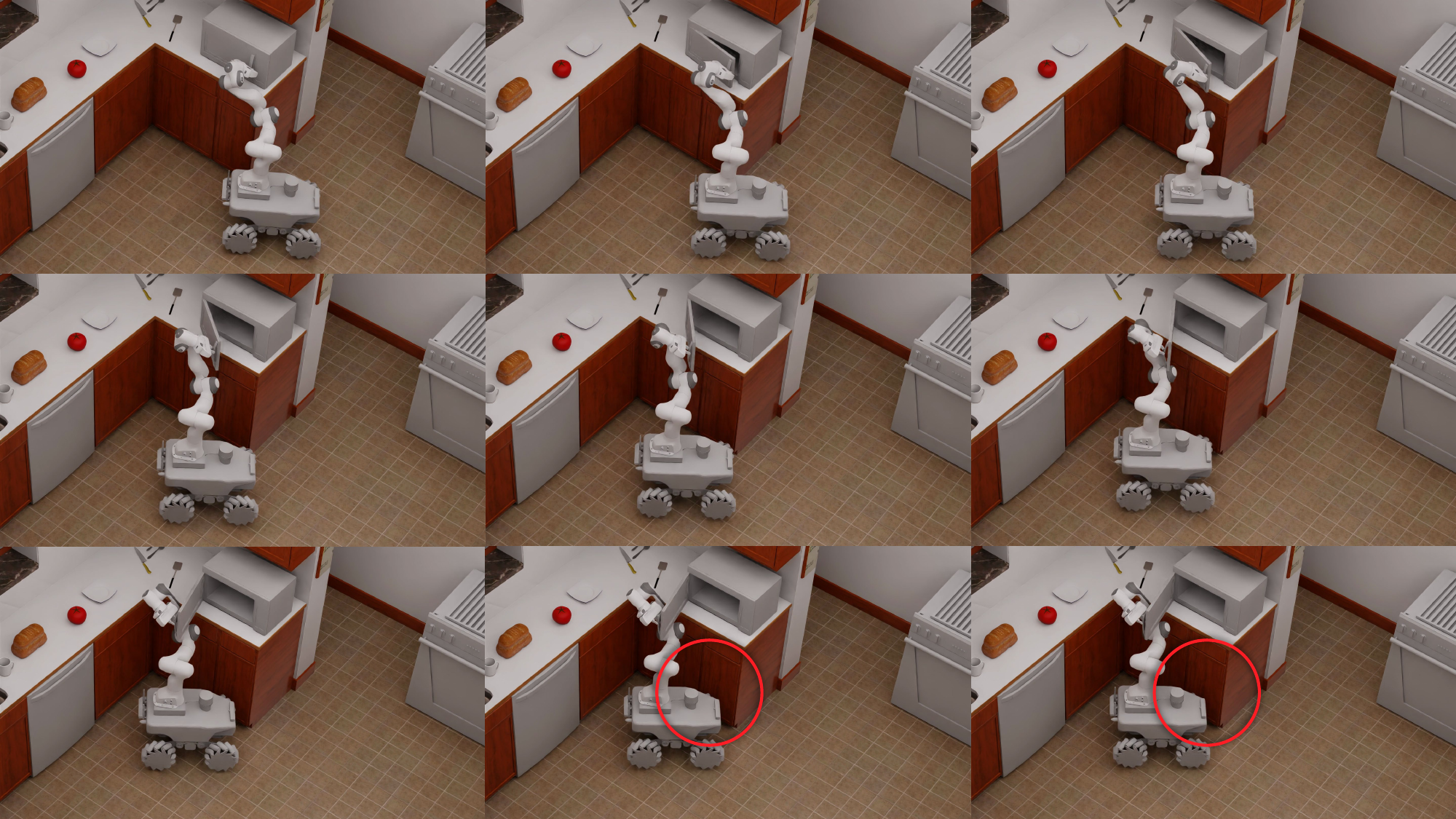}
    \caption{\textbf{Trajectory failure caused by collision.} Sphere-based geometry approximations occasionally fail to capture the original shape precisely, resulting in unintended collisions during planning.}
    \label{fig:appendix-fail-collision}
\end{figure*}

\paragraph{Failure due to collision}
\cref{fig:appendix-fail-collision} illustrates a collision failure. Sphere-based geometry approximations, while critical for \acs{gpu} acceleration, occasionally fail to capture the original shape precisely, causing unintended collisions during planning.

\begin{figure*}[t!]
    \centering
    \small
    \includegraphics[width=\linewidth]{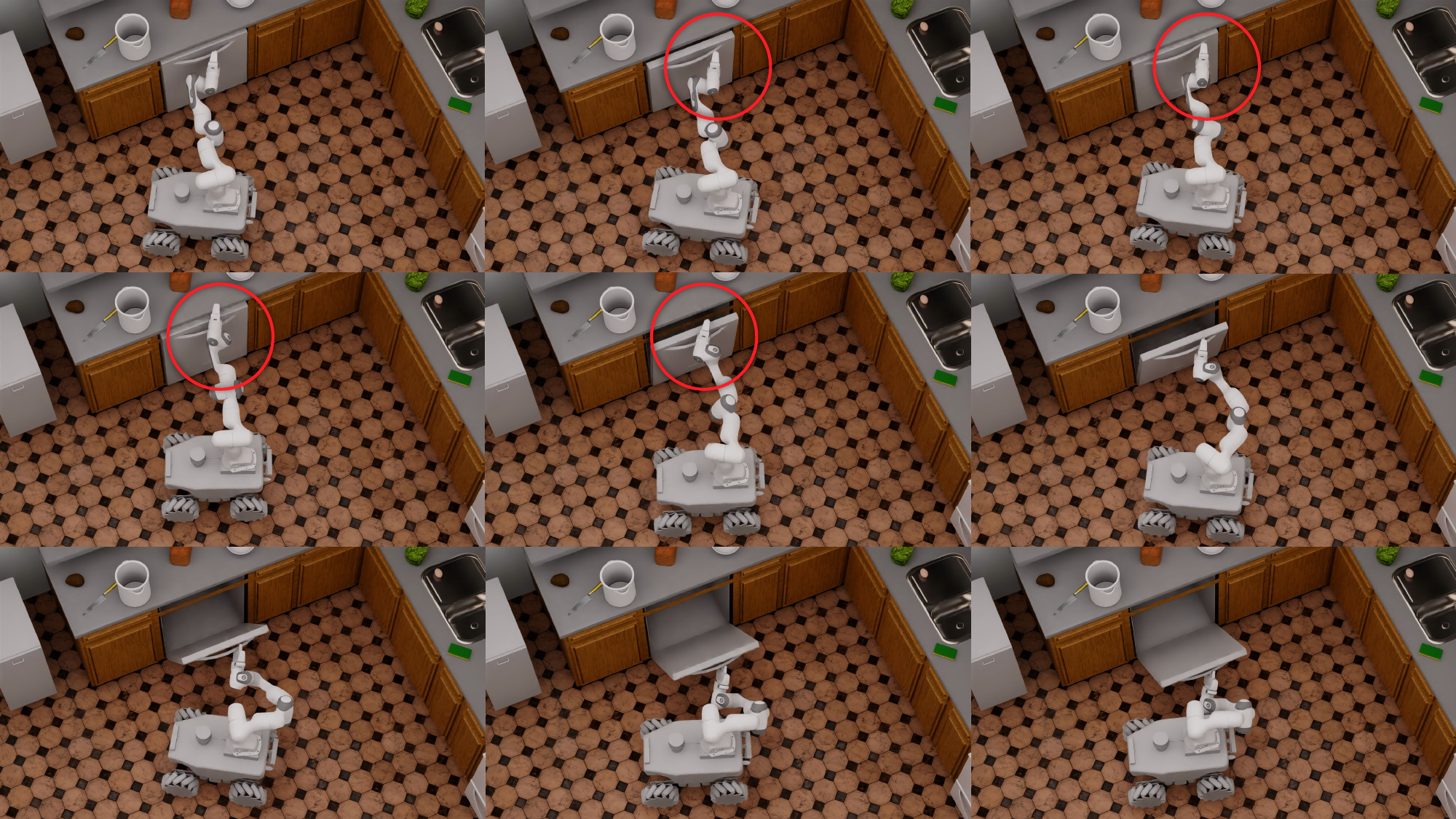}
    \caption{\textbf{Trajectory failure caused by fixed-base constraint violation.} The planned trajectory erroneously involves movements inconsistent with the object's fixed-base constraint.}
    \label{fig:appendix-fail-constraint}
\end{figure*}

\paragraph{Failure due to constraint violation}
\cref{fig:appendix-fail-constraint} depicts a fixed-base constraint violation, where the planned trajectory involves movements inconsistent with the object's environmental attachment. Ensuring strict constraint adherence remains challenging in complex manipulation scenarios.

\begin{figure*}[t!]
    \centering
    \small
    \includegraphics[width=\linewidth]{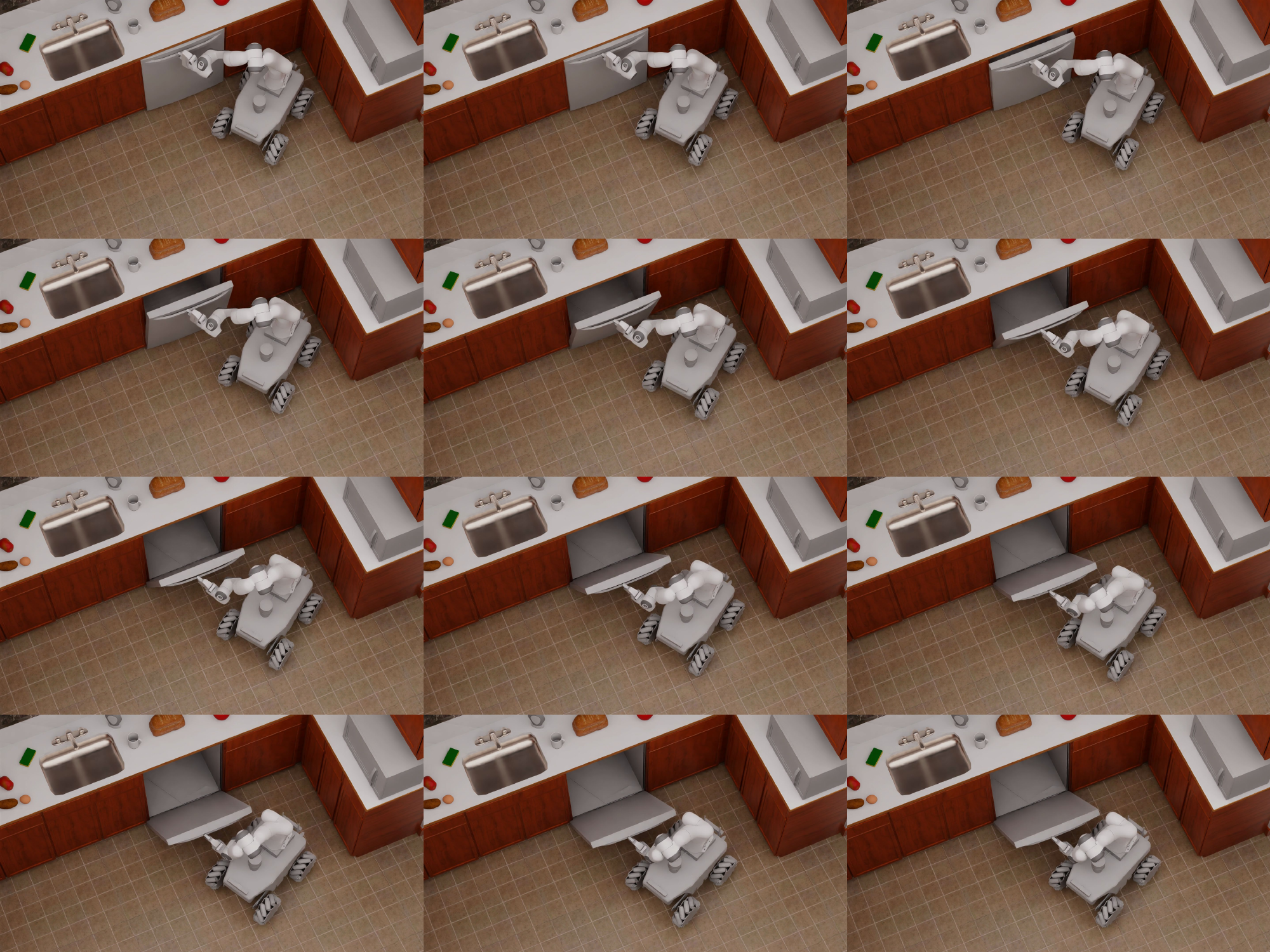}
    \caption{\textbf{Representative successful trajectory.} The planned trajectory correctly respects both collision and motion constraints, producing a physically plausible whole-body motion.}
    \label{fig:appendix-success-traj}
\end{figure*}

\paragraph{Successful trajectory}
\cref{fig:appendix-success-traj} illustrates a successful trajectory in which both collision and motion constraints are correctly respected, validating the pipeline's capacity to generate physically plausible whole-body motions across diverse tasks.

\subsection{Representative Inference Success and Failure Cases}\label{supp:sec:inference_cases}

Experiments are conducted on 30 training scenes and 10 held-out test scenes; the number of scenes is varied per experimental setting to ensure fair evaluation. The three success cases illustrate that with sufficient scale and diversity, the policy learns robust whole-body motion across varied layouts and object configurations. The failure case shows that errors arise not from misunderstanding articulation structure, but from pose inaccuracies that accumulate over execution.

\subsection{Real-World Validation}\label{supp:sec:real_world}

\begin{figure*}[t!]
    \centering
    \small
    \begin{subfigure}[b]{\linewidth}
        \centering
        \includegraphics[width=0.9\linewidth]{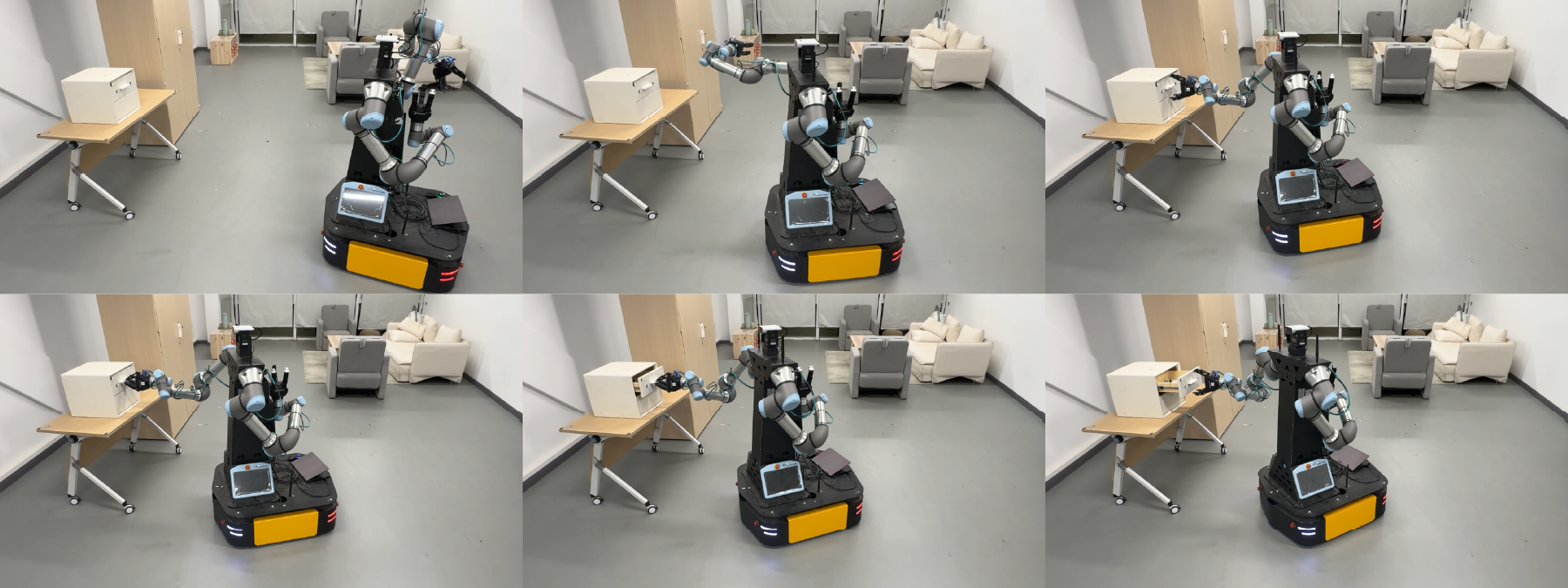}
        \caption{Drawer-opening trajectory on the UR5-Ridgeback platform.}
        \label{fig:appendix-real-drawer}
    \end{subfigure}%
    \\%
    \begin{subfigure}[b]{\linewidth}
        \centering
        \includegraphics[width=0.9\linewidth]{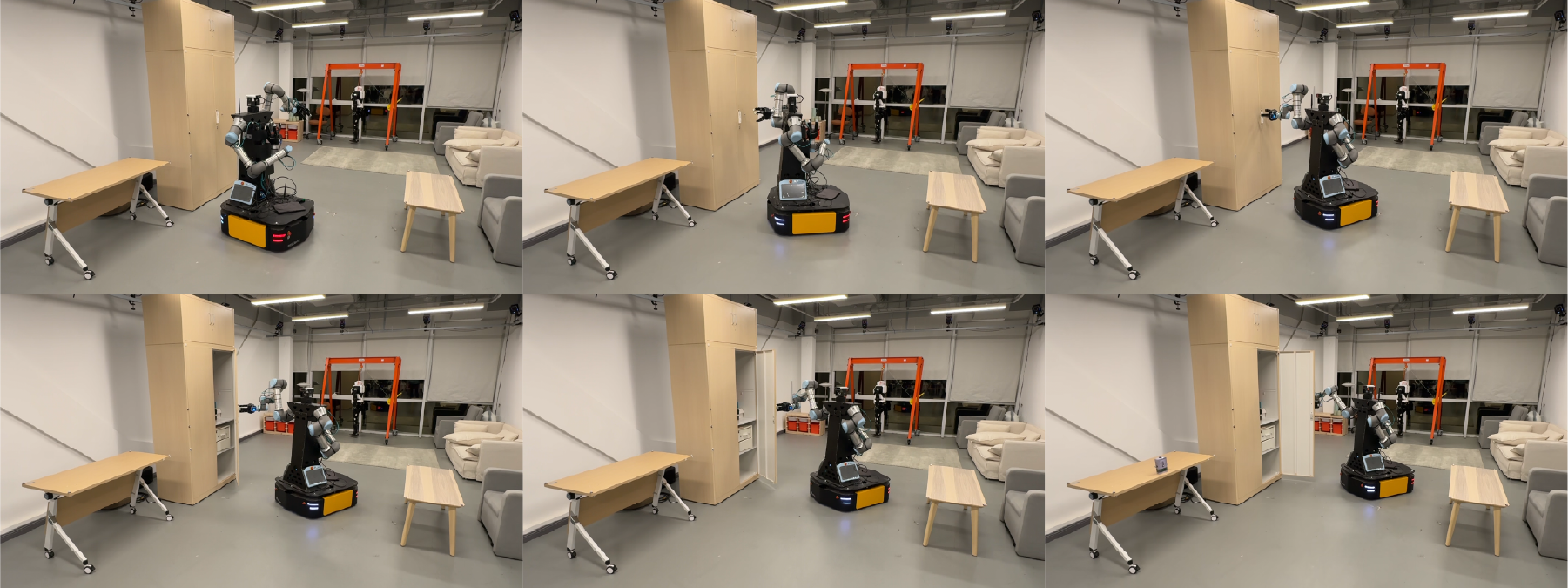}
        \caption{Cabinet door opening trajectory on the UR5-Ridgeback platform.}
        \label{fig:appendix-real-cabinet}
    \end{subfigure}%
    \caption{\textbf{Real-world validation on a UR5-Ridgeback platform.} Planned trajectories for drawer opening and cabinet door opening are executed smoothly without collision or constraint violation.}
    \label{fig:appendix-real-world-validation}
\end{figure*}

We validate the planning pipeline on a physical UR5-Ridgeback system comprising two UR5 manipulators mounted on a Clearpath Ridgeback mobile base. On both drawer opening and cabinet door opening tasks, the robot executes the planned trajectories smoothly, accurately reproducing simulation-generated motions without collisions or constraint violations (\cref{fig:appendix-real-world-validation}).

\section{Additional Ablation Experiments}

\subsection{Scaling Analysis up to 100k Trajectories}\label{supp:sec:scaling_100k}

To systematically analyze the effect of dataset scaling, we evaluate the DP3 policy trained on subsets of 10k, 30k, 50k, and 100k trajectories, spanning five SAPIEN objects across 30 scenes. \cref{fig:per_object_success_100k} details the per-object success rates under both seen and unseen configurations across these varied data scales. 

As the trajectory count increases, we observe a consistent upward trend in success rates for both seen and unseen settings across all evaluated objects. At the maximum scale of 100k trajectories, the policy effectively handles multiple object geometries, achieving over 50\% success on seen environments for four of the five objects. Notably, the performance on unseen configurations scales closely with the seen performance, demonstrating improved zero-shot generalization as data volume expands. 

Performance variance across object categories remains largely correlated with physical constraints. For instance, the consistently lower success rate for object ID 46197 across all data scales corresponds to its unique articulation limits, which restrict the robot's valid manipulation workspace. Overall, this scaling analysis indicates that increasing trajectory density systematically improves the policy's capacity to handle varied kinematic structures and generalize to unseen states.

\subsection{Impact of Unique Start States on Training Performance}

\begin{figure*}[t!]
    \centering
    \small
    \begin{subfigure}[b]{.5\linewidth}
        \centering
        \includegraphics[width=\linewidth]{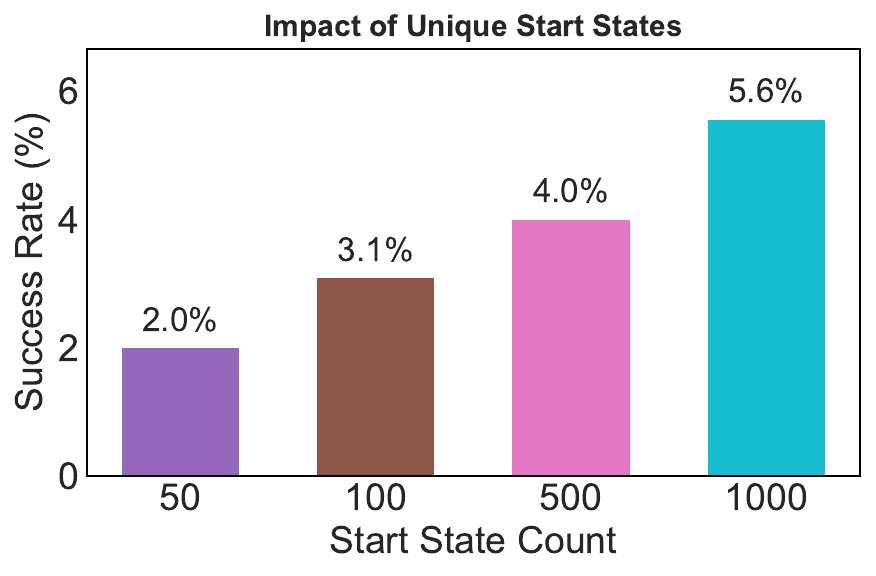}
        \label{fig:supp-ablation-startstate-1000}
    \end{subfigure}%
    \begin{subfigure}[b]{.5\linewidth}
        \centering
        \includegraphics[width=\linewidth]{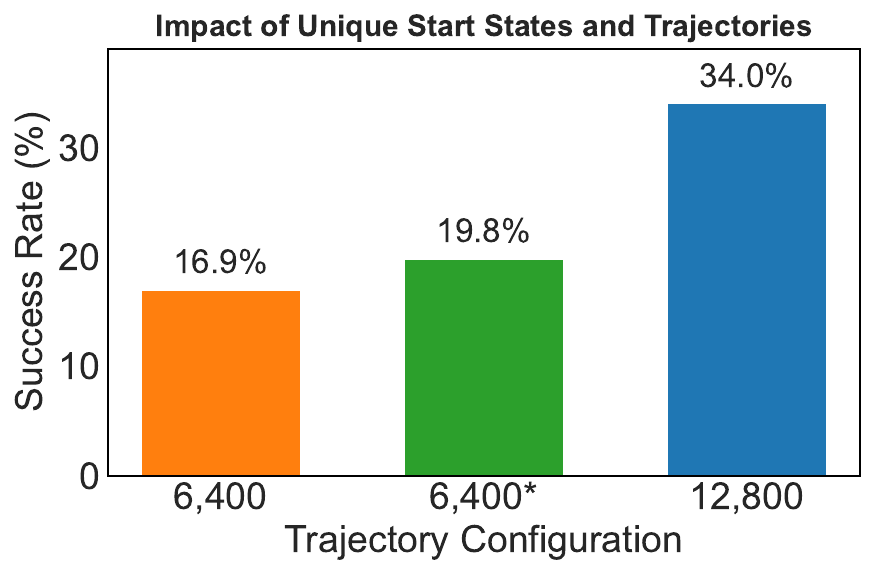}
        \label{fig:supp-ablation-startstate-6400-12800}
    \end{subfigure}%
    \caption{\textbf{Ablation on start-state diversity.} \textbf{Top:} Increasing unique start states (50--1,000) monotonically improves success on the 1,000-trajectory benchmark. \textbf{Bottom:} For fixed dataset sizes (6,400 trajectories), more diverse start states improve performance; the full 12,800-trajectory dataset achieves the highest success rate.}
    \label{fig:supp-ablation-startstate-all}
\end{figure*}

A central finding is that the number of unique start states has a direct and significant impact on policy generalization: even with a fixed trajectory count, increasing the diversity of initial configurations consistently improves downstream success rates (\cref{fig:supp-ablation-startstate-all}).

\paragraph{Effect of start-state diversity on the 1,000-trajectory benchmark}
Evaluating training sets with 50, 100, 500, and 1,000 unique start states on a fixed benchmark of 1,000 trajectories, a clear monotonic improvement emerges despite all configurations achieving a low absolute success rate of around 5\%. Broader coverage of initial base and arm configurations enables the diffusion policy to learn more reliable whole-body motion even under challenging test conditions.

\paragraph{Start-state diversity under fixed dataset sizes}
Using 6,400 trajectories, we compare a variant with 2,713 unique start states (raw distribution) against one with 5,244 unique start states (subsampled from 12,800 generated trajectories). The latter shows a clear improvement, demonstrating that start-state variety matters independently of dataset size. Training on the full 12,800 trajectories yields the highest performance, reflecting the combined benefit of higher data volume and richer start-state coverage.

\subsection{Pick Task Performance}\label{supp:sec:pick_ablation}

\begin{figure*}[t!]
    \centering
    \small
    \begin{subfigure}[b]{\linewidth}
        \centering
        \includegraphics[width=\linewidth]{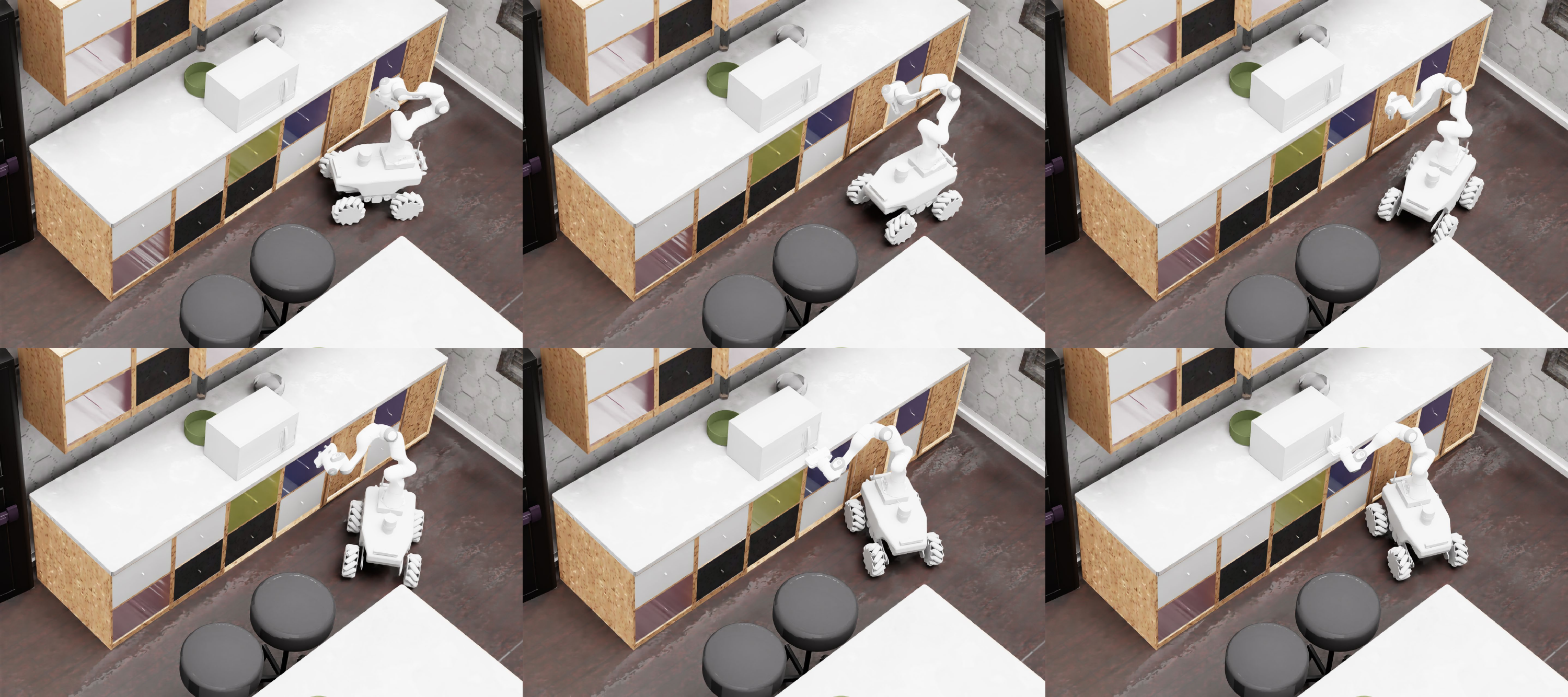}
        \caption{Successful execution with coordinated base-arm motion.}
        \label{fig:pick-success}
    \end{subfigure}%
    \\%
    \begin{subfigure}[b]{\linewidth}
        \centering
        \includegraphics[width=\linewidth]{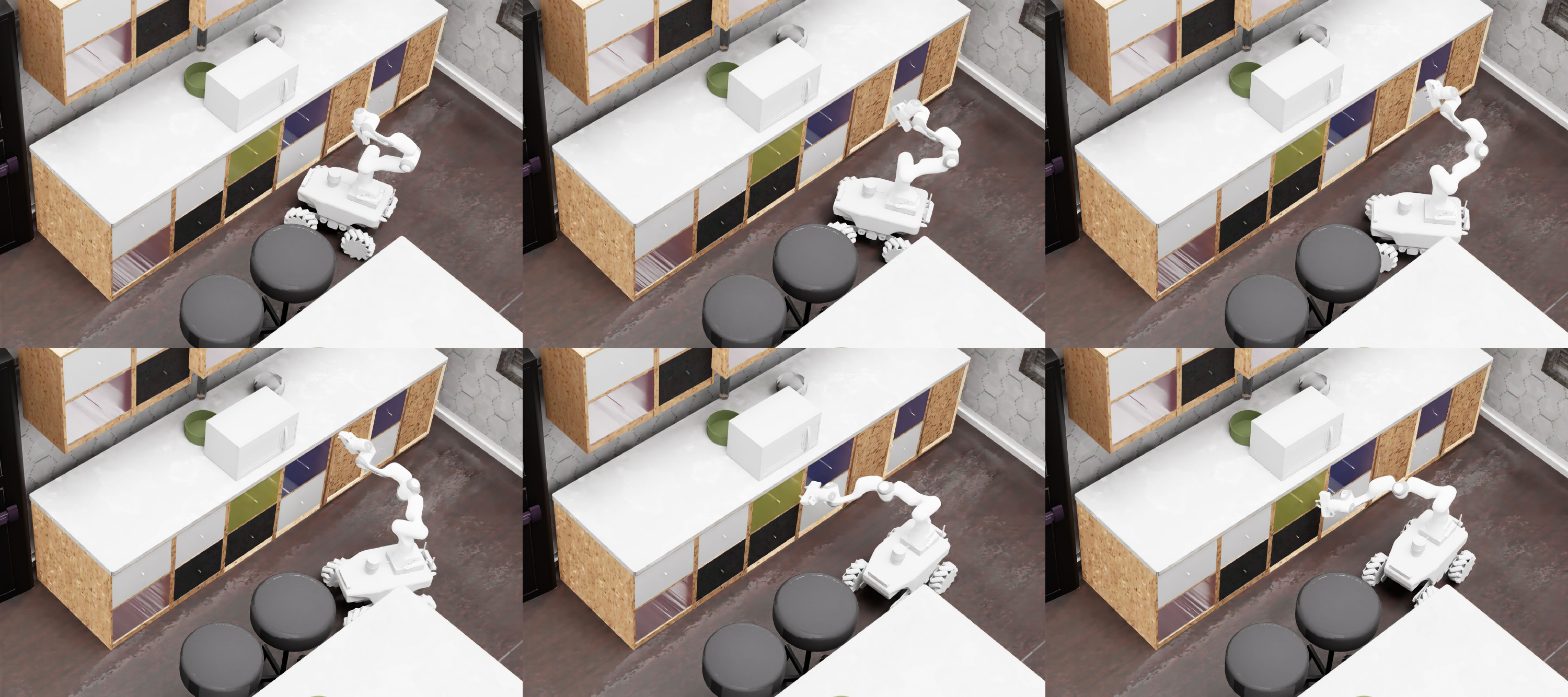}
        \caption{Failure case where the policy fails to stabilize the grasp.}
        \label{fig:pick-failure}
    \end{subfigure}%
    \caption{\textbf{Inference examples for the pick task} using a policy trained on 1,000 \dataset trajectories.}
    \label{fig:pick-inference}
\end{figure*}

We evaluate \dataset on a rigid-object Pick task by training a policy on 1,000 \dataset trajectories and comparing against a baseline trained on an equal amount of MoMaGen~\cite{li2025momagen} data. \dataset achieves a success rate of \textbf{51.92\%} at the reaching stage, outperforming the MoMaGen baseline at \textbf{35\%}, suggesting that physically valid whole-body motions are critical for sample-efficient policy learning. Representative inference results are shown in \cref{fig:pick-inference}.

\subsection{\texorpdfstring{Impact of Grasp-Switching (\vs Fixed-Grasp)}{Impact of Grasp-Switching (vs Fixed-Grasp)}}\label{supp:sec:grasp_switch}

\begin{figure*}[t!]
    \centering
    \small
    \begin{subfigure}[b]{\linewidth}
        \centering
        \includegraphics[width=\linewidth]{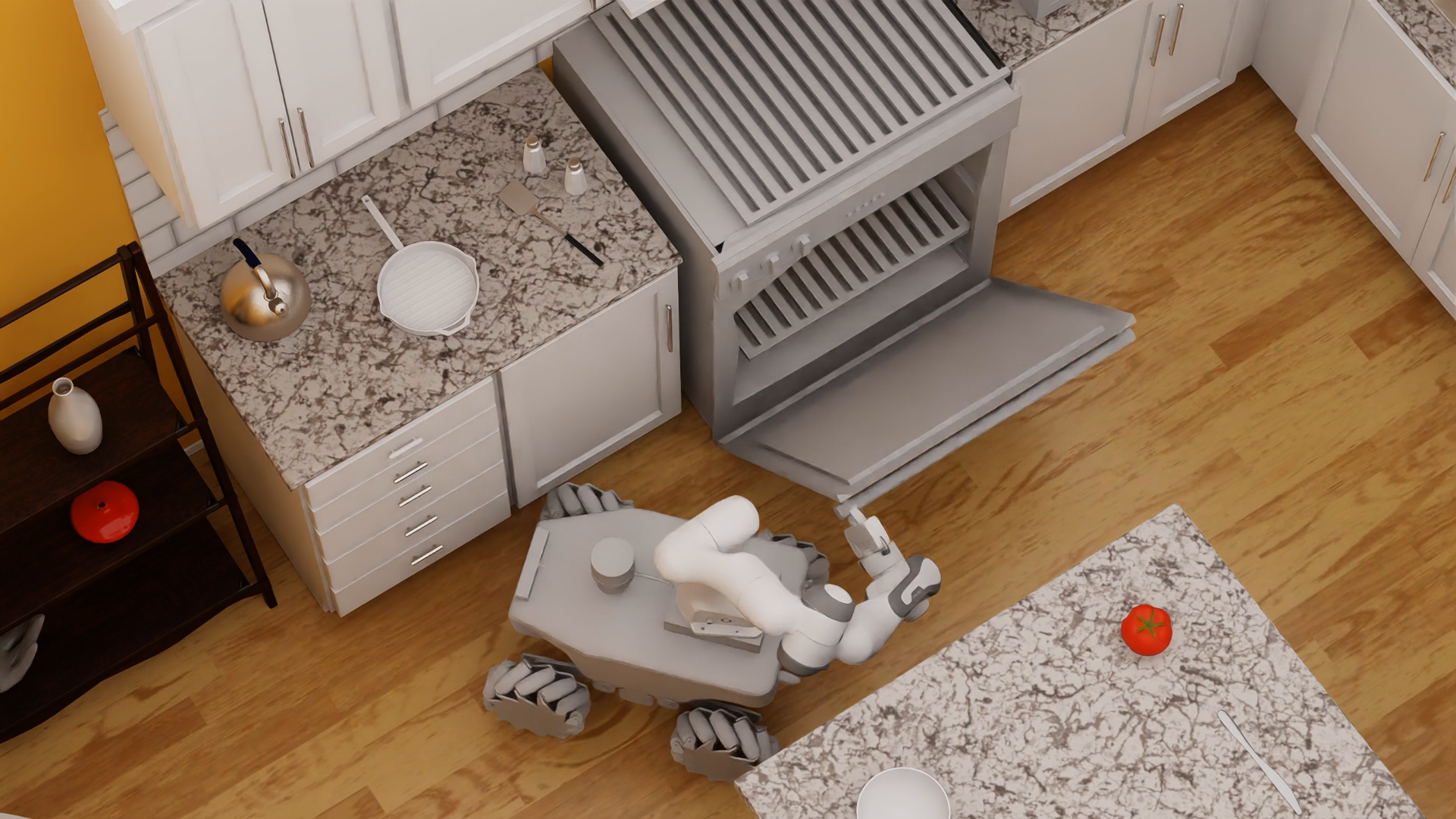}
        \caption{Fixed-grasp trajectory.}
        \label{fig:ablation-fixed}
    \end{subfigure}%
    \\%
    \begin{subfigure}[b]{\linewidth}
        \centering
        \includegraphics[width=\linewidth]{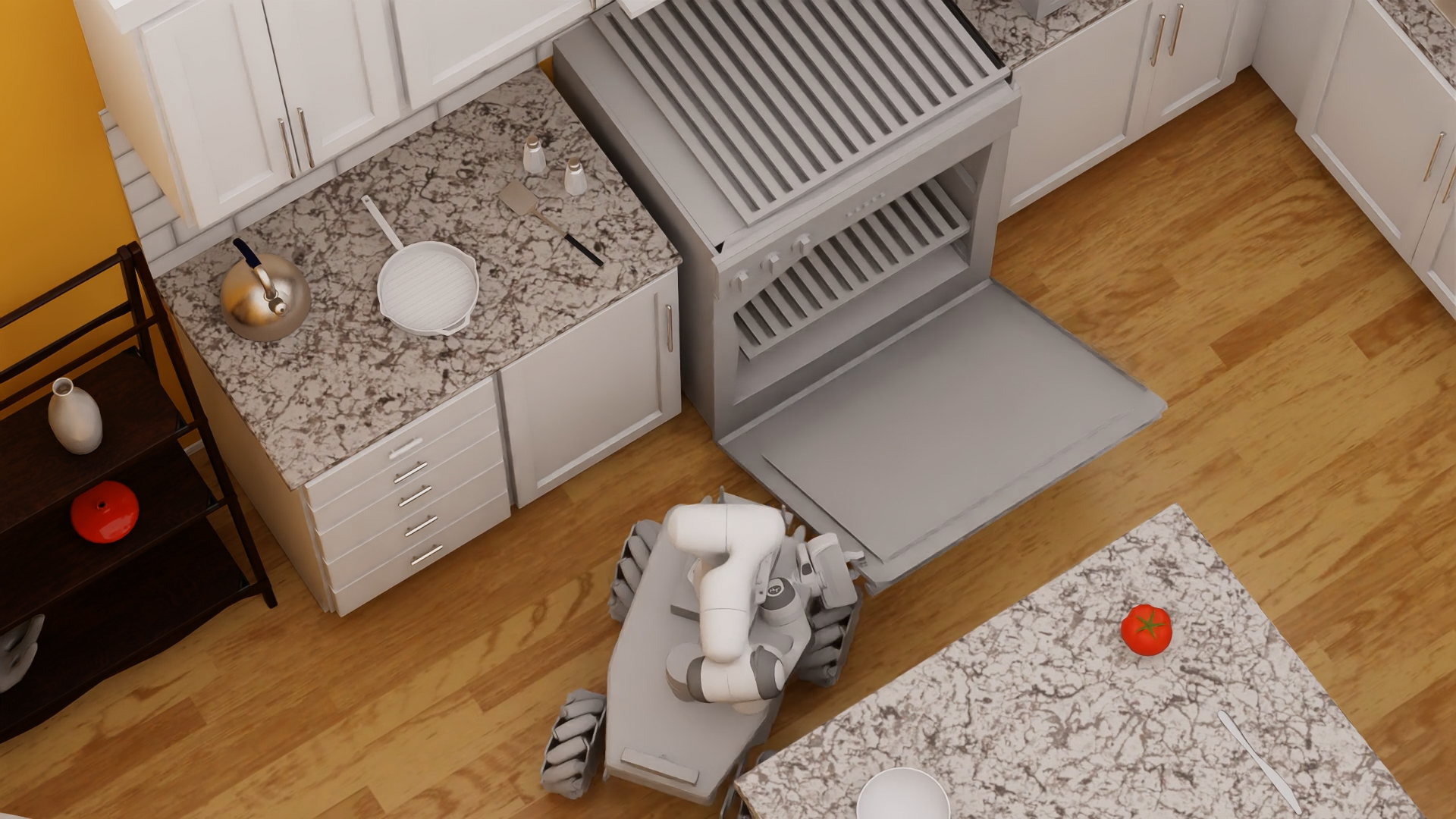}
        \caption{Grasp-switching trajectory.}
        \label{fig:ablation-switch}
    \end{subfigure}%
    \caption{\textbf{Fixed-grasp \vs grasp-switching.} Grasp switching enables larger object opening angles by avoiding link collisions.}
    \label{fig:ablation-fixed-switch}
\end{figure*}

We compare trajectories generated with a fixed grasp against those allowing grasp-switching when beneficial (\cref{fig:ablation-fixed-switch}). Grasp-switching enables the robot to change the grasp pose mid-task, increasing the achievable opening angle by avoiding link collisions. Trajectories with grasp-switching consistently attain larger maximum opening angles than fixed-grasp baselines, demonstrating its importance in constrained whole-body manipulation scenarios.

\begin{table}[ht]
    \centering
    \small
    \setlength{\tabcolsep}{3pt}
    \caption{Hyperparameters for \dataset data generation.}
    \label{supp:tab:automoma_config}
    \begin{tabular}{cc}
        \toprule
        Hyperparameter & Value \\ \midrule
        \multicolumn{2}{c}{\textbf{Environment and Collision}} \\ \midrule
        Object offset base (quaternion) & [0, 0, 0, 1, 0, 0, 0] \\
        Expanded dimension & 0.3 \\
        Disable collision & False \\
        Collision type & Voxel / Mesh \\
        \acs{ik} mesh pitch & 0.05 \\
        Trajectory mesh pitch & 0.05 \\
        Voxel dimensions & [5.0, 5.0, 5.0] \\
        Voxel size & 0.02 \\ \midrule
        \multicolumn{2}{c}{\textbf{IK and Trajectory Generation}} \\ \midrule
        Maximum retries & 10 \\
        $k$-means clusters & 500 \\
        \acs{ap} clusters upper bound & 80 \\
        \acs{ap} clusters lower bound & 10 \\
        \acs{ap} fallback clusters & 30 \\
        Position norm tolerance & 0.01 \\
        Rotation norm tolerance & 0.01 \\
        Grasp poses per object & 20 \\
        Open angles & 4--10 \\ \midrule
        \multicolumn{2}{c}{\textbf{Cameras and Sensing}} \\ \midrule
        Ego top-down camera & True \\
        Ego wrist camera & True \\
        Fixed camera & True \\
        Camera resolution & [320, 240] \\
        Camera frequency & 30\,Hz \\
        Focal length (ego top-down) & 50 \\
        Focal length (ego wrist) & 15 \\
        Focal length (fixed) & 50 \\
        Downsample point cloud & True \\
        Downsampled point count & 4096 \\
        \bottomrule
    \end{tabular}
\end{table}

\begin{table}[ht]
    \centering
    \small
    \setlength{\tabcolsep}{3pt}
    \caption{Hyperparameters for DP3~\cite{ze20243d}.}
    \label{tab:dp3_hyperparameter}
    \begin{tabular}{cc}
        \toprule
        Hyperparameter & Value    \\ \midrule
        \multicolumn{2}{c}{\textbf{Observation and Action}} \\ \midrule
        Number of points in point cloud & 4096 \\
        Point feature dimension & 3 \\
        Proprioceptive input dim & 11 \\
        History length (observation steps) & 3 \\
        Prediction horizon & 8 \\
        Number of action steps & 6 \\
        Action dim & 11 \\ \midrule
        \multicolumn{2}{c}{\textbf{Point Cloud Encoder}} \\ \midrule
        Use point crop & True \\
        Crop shape & [80, 80] \\
        PointNet input channels & 3 \\
        PointNet output dim & 64 \\
        PointNet uses LayerNorm & True \\
        PointNet final norm & LayerNorm \\
        Use normal channel & False \\
        Use point colors & False \\
        PointNet type & pointnet \\ \midrule
        \multicolumn{2}{c}{\textbf{Diffusion Backbone}} \\ \midrule
        Condition type & FiLM \\
        Use down/mid/up conditioning & True / True / True \\
        U-Net down dims  & [512, 1024, 2048] \\
        U-Net kernel size & 5 \\
        U-Net number of groups & 8 \\
        Diffusion step embedding dim & 64 \\
        Noise scheduler & DDIMScheduler \\
        Number of training diffusion steps & 100 \\
        Beta schedule & squaredcos\_cap\_v2 \\
        Prediction type & sample \\
        Denoise steps per inference & 10 \\ \midrule
        \multicolumn{2}{c}{\textbf{Optimization and Training}} \\ \midrule
        Optimizer & AdamW \\
        Learning rate & $1.0 \times 10^{-4}$ \\
        Adam betas & (0.95, 0.999) \\
        Adam $\epsilon$ & $1.0 \times 10^{-8}$ \\
        Weight decay & $1.0 \times 10^{-6}$ \\
        Batch size & 512 \\
        Number of data loader workers & 8 \\
        Number of epochs & 300 \\
        Learning rate scheduler & Cosine decay \\
        Learning rate warmup steps & 500 \\
        Gradient accumulation steps & 1 \\
        Use EMA & True \\
        EMA inv\_gamma & 1.0 \\
        EMA power & 0.75 \\
        EMA min / max value & 0.0 / 0.9999 \\
        Random seed & 42 \\
        \bottomrule
    \end{tabular}
\end{table}

\begin{table}[ht]
    \centering
    \small
    \setlength{\tabcolsep}{3pt}
    \caption{Hyperparameters for DP~\cite{chi2025diffusion}.}
    \label{tab:dp_hyperparameter}
    \begin{tabular}{cc}
        \toprule
        Hyperparameter & Value \\ \midrule
        \multicolumn{2}{c}{\textbf{Observation and Action}} \\ \midrule
        Number of camera views & 3 \\
        Proprioceptive input dim & 11 \\
        History length (observation steps) & 3 \\
        Prediction horizon & 8 \\
        Number of action steps & 6 \\
        Action dim & 11 \\
        Observation as global condition & True \\ \midrule
        \multicolumn{2}{c}{\textbf{Image Encoder}} \\ \midrule
        RGB backbone & ResNet-18 \\
        Pretrained weights & None \\
        Random crop augmentation & True \\
        Use group normalization & True \\
        Share RGB model across views & False \\
        ImageNet normalization & True \\ \midrule
        \multicolumn{2}{c}{\textbf{Diffusion Backbone}} \\ \midrule
        U-Net down dims & [256, 512, 1024] \\
        U-Net kernel size & 5 \\
        U-Net number of groups & 8 \\
        Diffusion step embedding dim & 128 \\
        Conditional predict scale & True \\
        Noise scheduler & DDPMScheduler \\
        Number of training diffusion steps & 100 \\
        Beta start / end & 0.0001 / 0.02 \\
        Beta schedule & squaredcos\_cap\_v2 \\
        Prediction type & epsilon \\
        Clip sample & True \\
        Denoise steps per inference & 100 \\ \midrule
        \multicolumn{2}{c}{\textbf{Optimization and Training}} \\ \midrule
        Optimizer & AdamW \\
        Learning rate & $1.0 \times 10^{-4}$ \\
        Adam betas & (0.95, 0.999) \\
        Adam $\epsilon$ & $1.0 \times 10^{-8}$ \\
        Weight decay & $1.0 \times 10^{-6}$ \\
        Batch size & 128 \\
        Number of data loader workers & 0 \\
        Number of epochs & 300 \\
        Learning rate scheduler & Cosine decay \\
        Learning rate warmup steps & 500 \\
        Gradient accumulation steps & 1 \\
        Mixed precision & fp16 \\
        Use EMA & True \\
        EMA inv\_gamma & 1.0 \\
        EMA power & 0.75 \\
        EMA min / max value & 0.0 / 0.9999 \\
        Random seed & 42 \\
        \bottomrule
    \end{tabular}
\end{table}

\begin{table}[ht]
    \centering
    \small
    \setlength{\tabcolsep}{3pt}
    \caption{Hyperparameters for ACT~\cite{fu2024mobile}.}
    \label{tab:act_hyperparameter}
    \begin{tabular}{cc}
        \toprule
        Hyperparameter & Value \\ \midrule
        \multicolumn{2}{c}{\textbf{Observation and Action}} \\ \midrule
        Number of camera views & 3 \\
        Action dim & 11 \\
        Action chunk size & 6 \\
        Temporal aggregation & False \\
        Mobile base mode & relative \\ \midrule
        \multicolumn{2}{c}{\textbf{Model Architecture}} \\ \midrule
        Image backbone & ResNet-18 \\
        Position embedding & sine \\
        Hidden dim & 512 \\
        Feedforward dim & 3200 \\
        Number of attention heads & 8 \\
        Encoder layers & 4 \\
        Decoder layers & 7 \\
        Dropout & 0.1 \\
        Pre-norm & False \\
        Dilation & False \\
        Masks & False \\ \midrule
        \multicolumn{2}{c}{\textbf{Optimization and Training}} \\ \midrule
        Optimizer & AdamW \\
        Learning rate & $1.0 \times 10^{-5}$ \\
        Backbone learning rate & $1.0 \times 10^{-5}$ \\
        Weight decay & $1.0 \times 10^{-4}$ \\
        KL weight & 1.0 \\
        Number of epochs & 2000 \\
        Number of episodes per task & 100 \\
        Episode length & 1000 \\
        Random seed & 0 \\
        \bottomrule
    \end{tabular}
\end{table}

\clearpage

\begin{figure*}[t!]
    \centering
    \small
    \begin{subfigure}[b]{.9\linewidth}
        \centering
        \includegraphics[width=\linewidth]{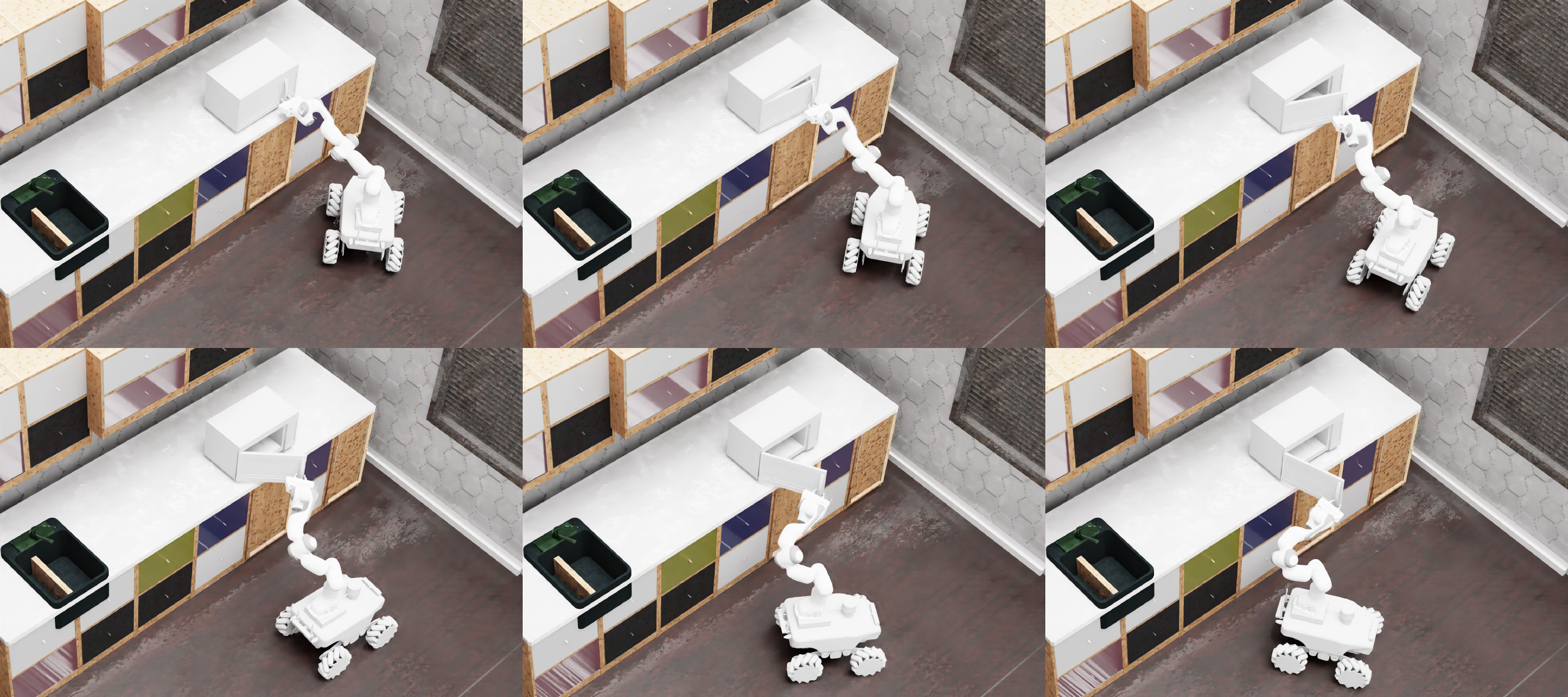}
        \label{fig:supp-inference-traj-success1}
    \end{subfigure}%
    \\\vspace{-1em}%
    \begin{subfigure}[b]{.9\linewidth}
        \centering
        \includegraphics[width=\linewidth]{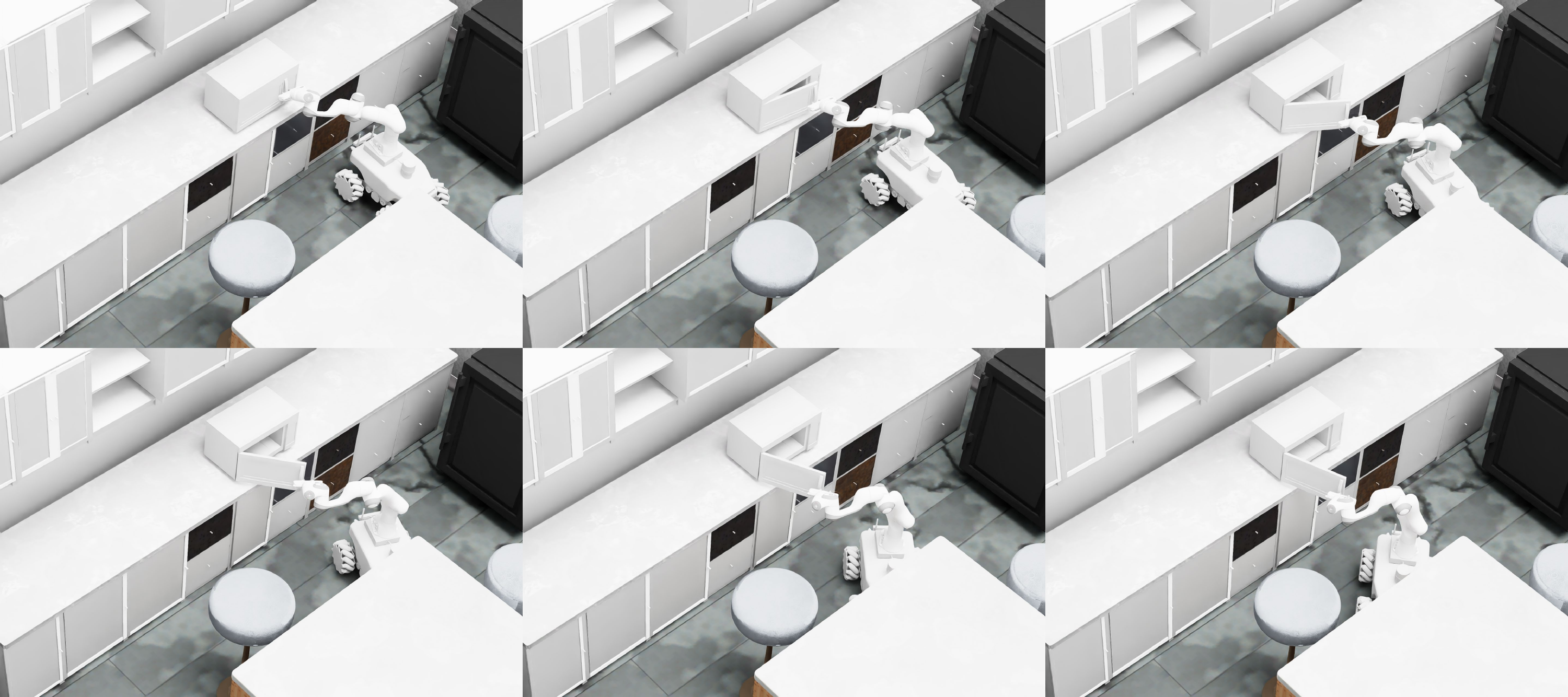}
        \label{fig:supp-inference-traj-success2}
    \end{subfigure}%
    \\\vspace{-1em}%
    \begin{subfigure}[b]{.9\linewidth}
        \centering
        \includegraphics[width=\linewidth]{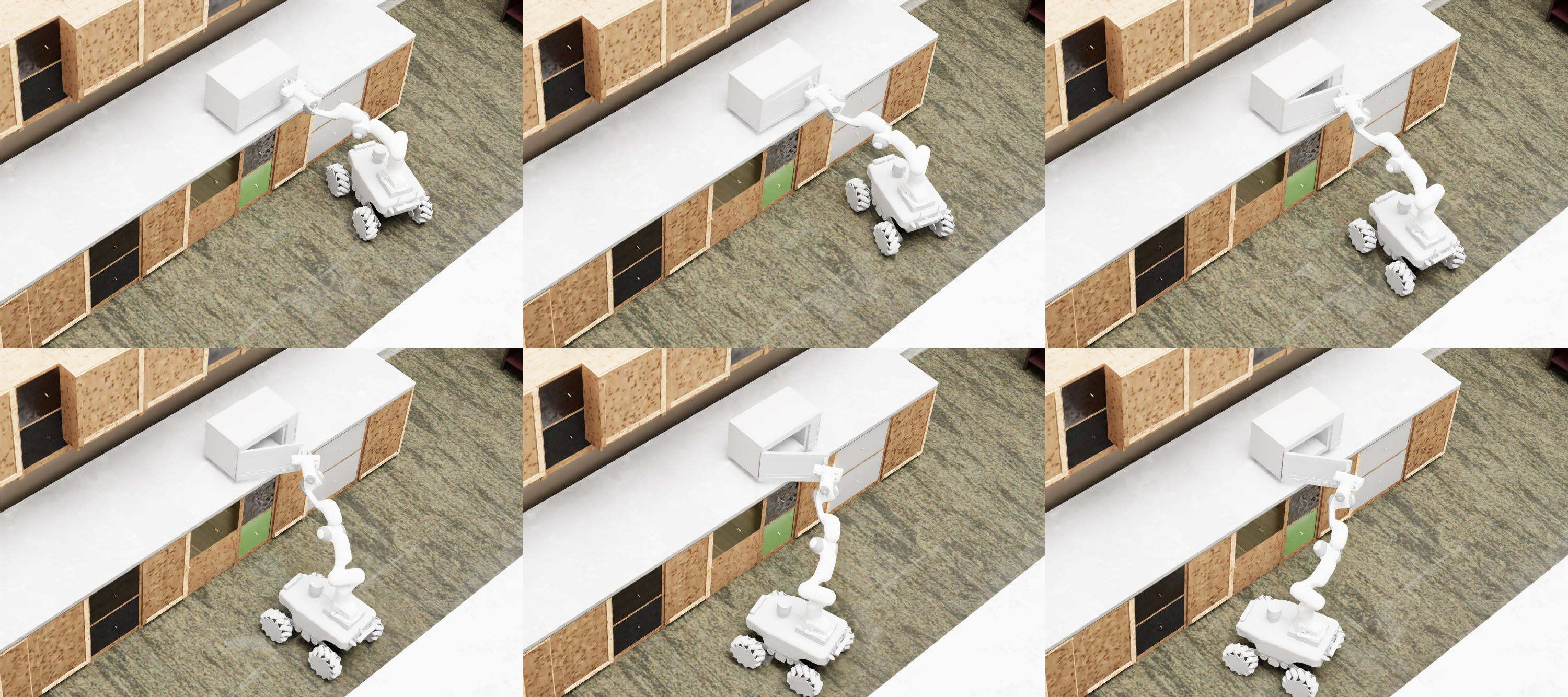}
        \label{fig:supp-inference-traj-success3}
    \end{subfigure}%
    \caption{\textbf{Representative successful inference trajectories across diverse scenes.} Trained on large-scale \dataset data, DP3~\cite{ze20243d} produces feasible whole-body trajectories across varied spatial layouts and articulated-object configurations.}
    \label{fig:supp-inference-traj-success-all}
\end{figure*}

\begin{figure*}[t!]
    \centering
    \small
    \includegraphics[width=0.9\linewidth]{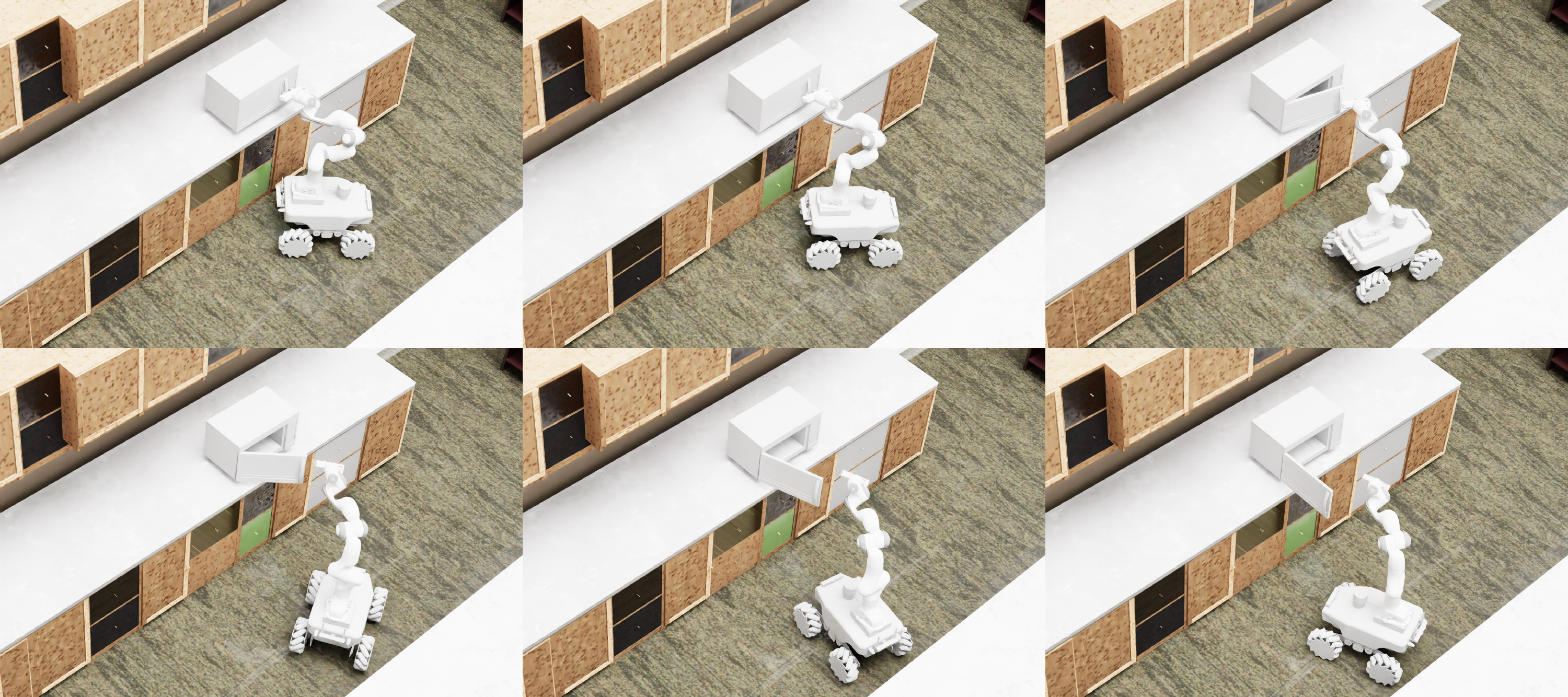}
    \caption{\textbf{Representative inference failure.} Small inconsistencies in base or arm pose predictions compound over time, eventually pushing the robot into an infeasible configuration.}
    \label{fig:supp-inference-traj-failure1}
\end{figure*}

\clearpage

\begin{figure*}[t!]
    \centering
    \small
    \includegraphics[trim=3mm 3mm 3mm 3mm,clip,width=\linewidth]{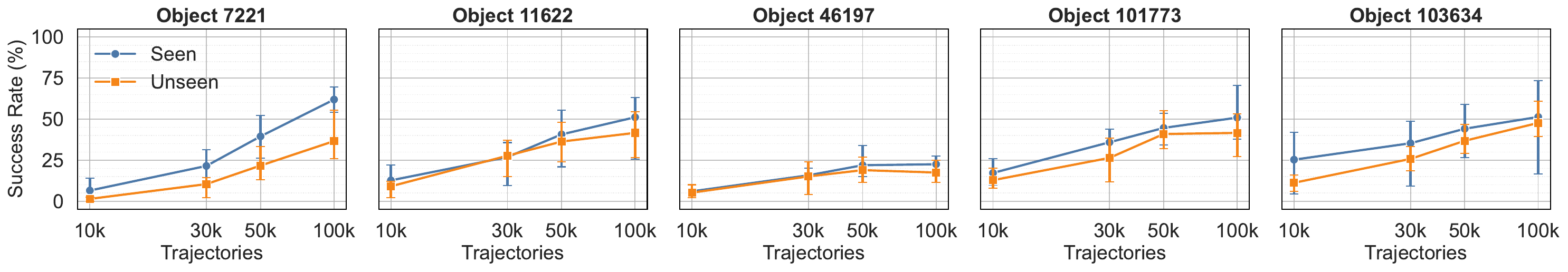}
    \caption{\textbf{Scaling analysis at 100k trajectories.} Training DP3 on a 100k-trajectory dataset spanning multiple articulated objects and scenes improves cross-object capability and generalization to unseen environments, compared with smaller-scale training subsets.}
    \label{fig:supp:scaling-100k}
\end{figure*}

\end{document}